\documentclass[journal]{IEEEtran}
\usepackage{capt-of}  
\usepackage{cuted}
\usepackage[backend=biber, sorting=none]{biblatex}
\usepackage{graphicx}%
\usepackage{multirow}%
\usepackage{amsmath,amssymb,amsfonts}%
\usepackage{amsthm}%
\usepackage{mathrsfs}%
\usepackage[figuresright]{rotating}%
\usepackage{xcolor}%
\usepackage{textcomp}%
\usepackage{manyfoot}%
\usepackage{booktabs}%
\usepackage{algorithm}%
\usepackage{algorithmicx}%
\usepackage{algpseudocode}%
\usepackage{program}%
\usepackage{listings}%
\usepackage{wrapfig}
\usepackage{vruler}
\usepackage{setspace}
\usepackage{comment}
\usepackage{siunitx}
\usepackage{booktabs}

\theoremstyle{thmstyleone}
\newtheorem{theorem}{Theorem}
\newtheorem{proposition}[theorem]{Proposition}
\theoremstyle{thmstyletwo}

\newtheorem{remark}{Remark}

\theoremstyle{thmstylethree}
\newtheorem{definition}{Definition}

\ifCLASSINFOpdf
\else
\fi
\begin{document}
%
\title{Computable Artificial General Intelligence}
%
%
%

\author{Michael~Timothy~Bennett
\thanks{M.T. Bennett was with the Australian National University e-mail: (see http://www.michaeltimothybennett.com).}
\thanks{Manuscript received 2022.}}

%
%

\markboth{preprint under review}%
{Shell \MakeLowercase{\textit{et al.}}: Bare Demo of IEEEtran.cls for IEEE Journals}
%



\maketitle

\begin{abstract}
Artificial general intelligence (AGI) may herald our extinction, according to AI safety research. Yet claims regarding AGI must rely upon mathematical formalisms -- theoretical agents we may analyse or attempt to build. AIXI appears to be the only such formalism supported by proof that its behaviour is optimal, a consequence of its use of compression as a proxy for intelligence. Unfortunately, AIXI is incomputable and claims regarding its behaviour highly subjective. We argue that this is because AIXI formalises cognition as taking place in isolation from the environment in which goals are pursued (Cartesian dualism). We propose an alternative, supported by proof and experiment, which overcomes these problems. Integrating research from cognitive science with AI, we formalise an enactive model of learning and reasoning to address the problem of subjectivity. This allows us to formulate a different proxy for intelligence, called weakness, which addresses the problem of incomputability. We prove optimal behaviour is attained when weakness is maximised. This proof is supplemented by experimental results comparing weakness and description length (the closest analogue to compression possible without reintroducing subjectivity). Weakness outperforms description length, suggesting it is a better proxy. Furthermore we show that, if cognition is enactive, then minimisation of description length is neither necessary nor sufficient to attain optimal performance. These results undermine the notion that compression is closely related to intelligence. We conclude with a discussion of limitations, implications and future research. There remain several open questions regarding the implementation of scale-able general intelligence. In the short term, these results may be best utilised to improve the performance of existing systems. For example, our results explain why Deepmind's Apperception Engine is able to generalise effectively, and how to replicate that performance by maximising weakness. Likewise in the context of neural networks, our results suggest both limitations of ``scale is all you need", and how those limitations can be overcome.
\end{abstract}


%
\IEEEpeerreviewmaketitle

\section{Introduction}
%
%
%
%
\label{intro}
\IEEEPARstart{W}{hat AGI} entails is debatable, but there does exist a mathematically precise definition -- an agent proven to maximise intelligence \cite{legg2007}\footnote{To clarify, maximising intelligence means first that intelligent behaviour is formally defined in measurable terms, and second that an agent behaves optimally according to that measure.}. Allowing for disagreement over what constitutes intelligence, AIXI is perhaps the only mathematical formalism proven to satisfy this criterion. It maximises ``the ability to satisfy goals in a wide range of environments" \cite{hutter2010,legg2008}. It is pareto optimal, meaning there is no agent which outperforms AIXI in one environment which can equal its performance in all others. 
AIXI is better able to learn and adapt, because the universal prior \cite{solomonoff_1964a,solomonoff_1964b} it employs allows it to make accurate inferences from minimal data. 
Unfortunately that universal prior is incomputable, ensuring AIXI can only ever be approximated. Nevertheless AIXI is a useful theoretical tool for understanding how an AGI may behave, for example in AI safety research \cite{cohen2022}. Yet even claims regarding AIXI's behaviour were later shown to be subjective, because behaviour depends upon the Universal Turing Machine (UTM) on which AIXI runs \cite{leike2015}. 
To address these problems, we propose an alternative.

\subsection{What is a universal prior?}
The source of AIXI's ``ability to satisfy goals in a wide range of environments" can be understood in simple terms.
All else being equal, the more accurately one predicts the future\footnote{To accurately predict the future means to infer which future among possible futures has the highest probability of occurring.} the better equipped one is to satisfy goals. AIXI is best able to satisfy goals because it can more accurately predict the consequences of its actions than any other agent. 
To predict what course of action will satisfy its goals, an agent needs a model of the context in which those goals are pursued. For AIXI, that model is a program able to perfectly reproduce what AIXI has experienced (a history of AIXI's interactions) \cite{hutter2010}. Given a history of interactions, there will exist many such models. Though these models all recreate the same past, they may differ in the future they predict. What makes AIXI intelligent is that it has a means of discerning which of these programs will predict the future accurately. It is named Kolmogorov Complexity (KC). The lower a model's Kolmogorov Complexity (KC) \cite{kolmogorov_1963}, the more likely it will correctly predict the future\footnote{This is a simplification. More formally, if the model which generated past data is indeed computable, then the simplest model will dominate the Bayesian posterior as more and more data is observed. Eventually, you will have identified the correct model and can use that model to generate the next sample (predict the future).}. 
This gives AIXI what is called a universal prior \cite{solomonoff_1964a,solomonoff_1964b}\footnote{To clarify, the universal prior AIXI employs was invented by Ray Solomonoff decades prior, albeit in the context of deterministic binary sequence prediction rather than reinforcement learning. It formalises Ockham's Razor, preferring simpler models where simplicity is estimated via Kolmogorov Complexity. However, Ockham's Razor is irrelevant in our context. Only the resulting behaviour matters.}. Given \textit{any} two models that are equivalent with respect to past events, a universal prior can determine which one is most likely to accurately predict the outcome of unforeseen future events. This is why AIXI is also called a \textit{universal} artificial intelligence \cite{goertzel2014}.
A universal prior is what distinguishes AIXI from agents not considered formalisms of AGI. It is what enables AIXI to maximise intelligence.

\subsection{Incomputability}
If two models are equivalent with respect to the past, then Kolmogorov Complexity (KC) can be used to determine which will best predict the future -- which of the two is a more plausible model of reality.
Because of this, AIXI can estimate one thing (how close a model is to reality) by looking at another, seemingly unrelated thing (KC). KC acts as a proxy for future predictive accuracy.
The KC of data is the shortest possible program which can reproduce that data, in the context of given UTM. Because KC is the length of the \textit{most compressed} representation of data, an agent that relies upon KC must be able to compress at least as well as any other agent. 
The point (in greatly simplified terms) is that AIXI maximises intelligence by maximising the ability to compress information.
Hence, compression is considered by some to be a proxy for intelligence \cite{chaitin2006}.
Unfortunately, the optimal compressed representation of something is incomputable, which makes KC incomputable. AIXI's intelligence hinges upon KC. It follows that AIXI is incomputable is because its very foundation is compression as a proxy for intelligence. Yes, there do exist compelling work-arounds and approximations \cite{orallo2010,leike2018}, but this association of compression and intelligence has gone largely unchallenged.

\subsubsection{Solution to incomputability}
A formalism of AGI that \textit{is} computable needs a computable universal prior. To get a computable universal prior, we need to identify a different proxy for intelligence -- a different way of discerning which model (among those that explain the past) will make the most accurate predictions in future.

\subsection{Subjectivity}
As mentioned earlier KC is measured in the context of the UTM on which AIXI runs. By itself, changing the UTM would not meaningfully affect performance. When used in a universal prior to predict deterministic binary sequences, the number of incorrect predictions a model will make is bounded by a multiple of the KC of that model \cite{solomonoff1978}. If the UTM is changed the number of errors only changes by a constant  \cite[2.1.1 \& 3.1.1]{vitanyi2008}, so changing the UTM doesn't change which model is considered most plausible. However, when AIXI employs this prior in an \textit{interactive} setting, a problem occurs \cite{leike2015}. To explain in simplified terms (with some abuse of notation), assume a program $f_1$ is an agent's mind, $f_2$ is an interpreter (a UTM) and $f_3$ is the reality (physical body and environment) within which goals are pursued. Intelligence is a measure of the performance of $f_3(f_2(f_1))$. AIXI is the optimal choice of $f_1$ to maximise the performance of $f_3(f_2(f_1))$. However, in an interactive setting the perception of success may not match reality. 

\begin{quote}
``Legg-Hutter intelligence \cite{legg2007} is measured with respect to a fixed UTM. AIXI is the most intelligent policy if it uses the same UTM." \cite[p.10]{leike2015}
\end{quote}

\noindent If intelligence is measured with respect to one UTM while AIXI runs on another, then this is like AIXI being engaged in one reality, while success is determined by another, entirely different reality. $f_3(f_2(f_1))$ depends upon $f_2(f_1)$, not $f_1$ alone. Thus the performance of $f_1$ alone is considered to be \textit{subjective}. 

\begin{quote}
``This undermines all existing optimality properties for AIXI." \cite[p.1]{leike2015}
\end{quote}

\subsubsection{Solution to subjectivity}
To measure performance objectively, we would need to remove the interpreting interface between the agent and reality.
Fortunately, this is a new incarnation of an old problem. Cartesian dualism \cite{howard2020} is the philosophical notion that mind and body are separate. AIXI seems to formalise Cartesian dualism, in that the mind $f_1$ is only connected to reality $f_3$ via a UTM $f_2$.
In other fields, dualism has fallen out of favour because there is far more evidence to support the idea that cognition is embodied, extends into the environment and is enacted by organisms within a purely physical world \cite{shapiro2021}. For example, if someone uses pen and paper to compute the solution to a math problem, then their cognition is extending into and enacted within the environment \cite{clark1998}. Cognition occurs \textit{within} the reality that determines what is true and what behaviour will be successful. Because of this, a scenario where intelligence is measured with respect to one reality while cognition takes place in another cannot arise. Formalising enactive cognition \cite{ward2017} would allow us to eliminate the unnecessary distinction between thoughts and actions in favour of general sensorimotor activity. What matters is whether sensorimotor activity satisfies goals \cite{bennett2021}, not the distinction between computation and actuation. Notions like mind, body and environment can be discarded -- what we have instead is a system of sensiromotor circuitry through which sensorimotor activity is enacted. 
To reconcile enactivism with computation, it would need to be integrated with pancomputationalism (the notion that everything is a computing system) \cite{piccinini2021}. This would make the universe the UTM, and all of reality (mind, body and environment) the software. The consequence is that where AIXI formalises $f_3(f_2(f_1))$, enactivism would formalise $f_2(f_3(f_1))$ which would allow us to maximise performance in terms of $f_3(f_1)$ without worrying about $f_2$. 
The unavoidable downside is that we must formalise cognition in a way that deviates from convention.
There can be no interpreting interface between $f_3$ and $f_1$, so we must formalise reality and then cognition directly within it.

\subsection{Approach}
\label{approach}
Cognition needs to be \textit{part} of reality. A simple way to formalise one thing as part of another is to define the former as a subset of the latter. According to the Curry-Howard isomorphism, it does not matter if we use a set of declarative or imperative programs \cite{howard1980} (describing what and how is equivalent if what is sufficiently detailed). Thus for simplicity we represent reality as a set of declarative programs, of which sensorimotor activity is a subset\footnote{$f_3$ is formally defined as the set of declarative programs $P$ in section \ref{definitions} definition \ref{d1}, $f_2$ is a model ${m} \in {M}$ in definition \ref{d3}}. To illustrate using the notation from our earlier example (with further abuse of notation), this means $f_1 \subset f_3$. The interpreter $f_2$ is the pancomputationalist's universe, giving us $f_2(f_3)$. From a computer science point of view this might be understood as there existing only one interpreter on which all programs are run. This addresses the problem of subjective performance because there is no interpreter between $f_1$ and $f_3$, however the question is now ``how do we formalise learning and reasoning" in this context?

To ``achieve goals in a wide range of environments" implies an agent's goals and model of the environment are independent. This is an unnecessary complication. 
The only parts of reality that actually matter to an agent are those pertaining to its goals. 

A task is a goal completed within the environment, using one's body and any other parts of the environment that might serve as tools. Intuitively this integrates a goal with the context in which it is pursued. It is subsequently a good basis for an enactive model of cognition. Where a model of an environment would describe what will happen in that environment, a model in the context of a task only needs to describe how to achieve a specific goal while embodied and embedded in a specific local environment (anything which is not necessary to satisfy the goal can be ignored) \cite{bennett2021}. The goal is implied by which aspects of the environment are modelled. Hence we do not need a model of the environment, but a model of a task \cite{bennett2021}.

Being part of reality, a task model is a set of declarative programs (because we described reality as a set of declarative programs). It may be thought of as the rules of the task in the sense that declarative programs are, intuitively, rules. Like AIXI's models, these task models are equivalent with respect to the past, but differ in what futures they predict. 
In the context of tasks ``the ability to satisfy goals in a wide range of environments" becomes ``the ability to complete a wide range of tasks". However, a model can describe how to complete more than one task. If one model describes how to complete a superset of the tasks another does, then the former can be used to complete a wider range of tasks than the latter. The number of tasks which a model can be used to complete is indicated by what is called its ``weakness". This replaces compression as our proxy for intelligence, and it is objective because there is no interpreting interface in the way. 
If the number of declarative programs (the aforementioned sensorimotor circuitry) used to describe the task is finite (a reasonable assumption for physical machinery), then we can compute which models (sets of declarative programs) entail completing the widest variety of tasks, and so have a computable prior based upon an objective proxy for intelligence. 

\subsubsection{Paper summary}
The key novel contributions of this paper are:
\begin{enumerate}
    \item A formalisation of enactive cognition, useful because it addresses the problems inherent in formalisations of dualism (e.g. that AIXI's claimed optimality is subjective).
    \item Formalisation of ``weakness" as a proxy for intelligence. In the context of task models (sets of declarative programs), the semantic notion of an extension and its accompanying intensions are opposite extremes of a spectrum ordered by weakness\cite{bennett2021,bennettmaruyama2021a}, so this mathematical definition can be used to measure if one model is ``more" of an intension than another. In the context of result 1 above, this is useful because an agent can \textit{learn} by finding weaker representations, extracting intensions (weak models) from examples of an extension (decisions).
    \item Formal proof that maximising weakness is necessary and sufficient to maximise intelligence in this context (result 1 above), which is useful as a computable, objective alternative to description length or compression. Weakness may serve to optimise the performance of existing methods, such as inductive logic programming (ILP) \cite{plotkin1972}.
    \item A computable prior based upon weakness, which is universal in the sense that it assigns belief to every concievable hypothesis. It is useful because it opens both theoretical and experimental avenues for future AGI research.
    \item Formal proof that minimising description length is not necessary to maximise intelligence in this context, while maximising weakness is.
    \item Experimental results comparing description length and weakness as proxies for intelligence. Our results show that weakness substantially outperforms minimum description length, suggesting 
    that weakness is a better proxy.
\end{enumerate}

We begin by walking through the some of the assumptions and background. Section \ref{definitions} formalises these notions mathematically, and proves the optimality of an agent that prefers weaker representations. These proofs are further supported by experimental evidence, given in section \ref{experiments}. The paper concludes with a discussion of limitations, and implications as they pertain to inductive logic programming (Deepmind's Apperception Engine \cite{evans_2020a,evans_2020b,evans_2021b} in particular), neural networks (scaling laws in particular) and the varying utility of general intelligence (intelligence is not always useful). The appendix gives examples, answers to frequently asked questions (FAQ) and additional definitions pertaining to related work.

\section{Premises}
\label{review}
This section provides some additional context for the formal definitions given in section \ref{definitions}.

\subsection{Modified agent-environment paradigm} 
\begin{figure}[t]
\centering 
\includegraphics[width=\columnwidth]{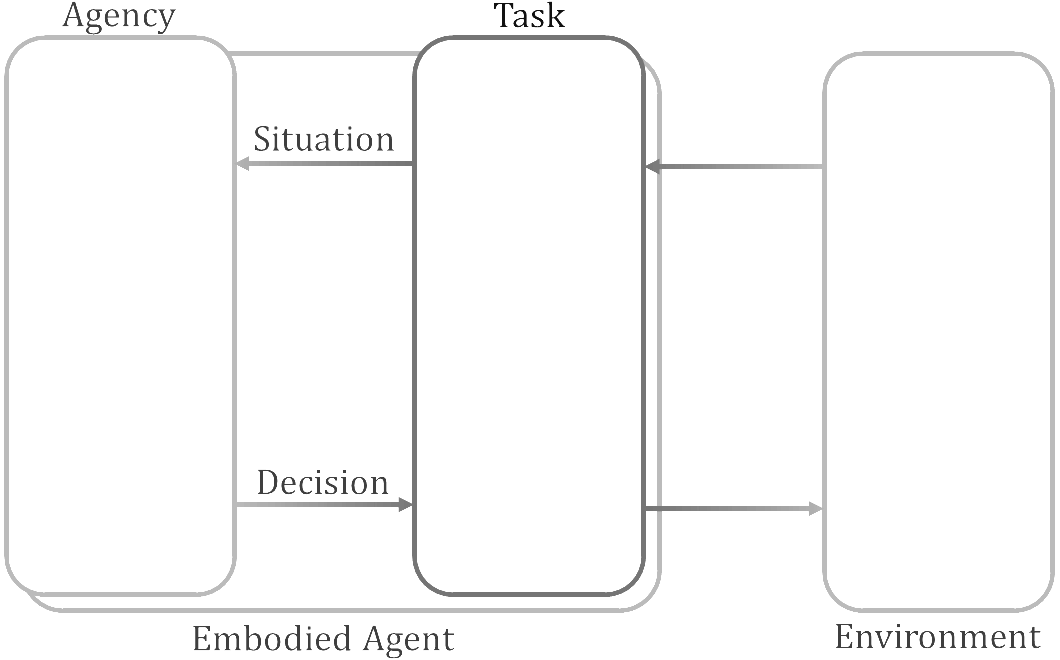}
\caption{Modified agent environment paradigm}.
\label{fig:1}
\end{figure}

Cognition is often understood using the agent-environment paradigm \cite{russellandnorvig2021}, which we have modified to account for tasks (see figure \ref{fig:1}).
\label{agi}
Enactivism \cite{ward2017,thompson2007} assumes an agent is embodied and embedded in an environment \cite{shapiro2021}, and cognition is enacted through that environment, extending into it. As mentioned in section \ref{intro} this prohibits the use of an interpreter between the mind and environment. This means there isn't necessarily a clear distinction between the agent and environment and so, instead of formalising agent and environment as separate objects, we need only formalise the task. After all, the tasks one engages in are an expression of both one's circumstances and agency (meaning ability to alter one's decisions in service of a goal when circumstances change \cite{kenton2022}). We only need to define the task (not agent and environment separately), because a completing a task requires completing the entire ``stack", from idea through execution. Enactive cognition eliminates the need to distinguish between thoughts and actions. A decision is just sensorimotor activity the agent engages in, in an attempt to complete the task.

\subsection{A single decision instead of sequential decisions} \label{seq}We are concerned with intelligent behaviour. Typically, models of intelligent behaviour assume behaviour is a sequence of actions or decisions over time (for example, AIXI). However, this assumption is unnecessary and limiting. Not all behaviour is sequences of actions. To formalise enactive cognition in general, such assumptions should be minimised. 

First, we discard the notion of discrete actions. Intuitively, an action is the result of a decision, and all intelligent behaviour involves decisions, but not all decisions result in action. Second, discard the idea of a sequence. It is common to represent planning problems (choose actions one after another in sequence) as boolean satisfiability problems (assign values to a set of variables in any order or all at once) \cite{kautz92}. This reframes a sequential decision problem as one decision, in the same way we might translate a an imperative program into a declarative program. All problems to which we might apply intelligence may be re-framed in terms of a solitary decision \cite{bennettmaruyama2021b}. A decision might set into motion a sequence of discrete actions, a continuous process, classify an object or take place in the middle of a whatever a previous decision set in motion. From a purely pragmatic point of view, what \textit{matters} if we want to define intelligence is not the structure of interaction between an agent and its environment, but the quality of a decision (does it lead to a desirable outcome?). Therefore we should go with the simplest structure possible, which is a single decision in the context of a task. Either that decision is better than the alternatives, or it is not. The resulting formalism can describe problems that involve sequences of decisions, such as MDPs (in the same way a plan can encode a sequence of actions), but it doesn't presuppose such an underlying structure.

\subsection{Binary classification instead of reward}
\label{binary}
Primitives of cognition such as ``pleasure" and ``hunger" are assumed because the agent is embodied. These primitive urges determine whether experiences are good or bad\footnote{This is not an endorsement of naive utilitarianism. We are simply stating that such urges do exist, and pointing out that they inform preferences.}. Subsequently they also determine whether the decisions to which those experiences are attributed are eventually considered good or bad\footnote{How experiences are attributed to past decisions is beyond the scope of this paper.}, imposing an order of preference on decisions. This ordering is akin to preference based reinforcement learning \cite{wirth2017}, which allows adaptation to ``non-Archimedean tasks whose rewards are too rich to express using real numbers" \cite{alexander2020}. For simplicity, we assume there exists a threshold beyond which a decisions are ``good enough", and that all past decisions are eventually labelled as either ``good enough" or ``not good enough" by some process of attribution (the details of which are beyond the scope of this paper). In other words, instead of a real valued reward signal (as is common in reinforcement learning), we frame cognition as a binary classification problem. In the context of a task, ``good enough" means the task is caused to become complete \cite{bennett2021,bennettmaruyama2021a}. 

\subsection{Accruing experience} 
A model is inferred from examples. Each example is a decision made in a situation. A set of examples is an ostensive definition \cite{gupta2021} or training set \footnote{An ostensive definition of a concept is a set of examples of that concept (positive examples). It does not need to contain examples of what the concept is not (negative examples). In this case, the concept is a task and each example is a decision together with the situation in which it was made. As learning from only positive examples is harder than learning from both positive and negative examples, this paper assumes only positive examples are given in ostensive definitions. However, models constructed from both positive and negative examples are formalised in the appendix and discussed in related work.}.

This paper assumes an ostensive definition is given, so that different models may be compared given the same ostensive definition. However, an ostensive definition may also be constructed through repeated task completion, like a reinforcement learning agent \cite{bennettmaruyama2021b} (assuming a means of attributing task completion to decisions)\footnote{One's present decisions would eventually become part of future ostensive definitions.}. As discusssed in section \ref{seq} sequential interactions can take place, but we don't need to explicitly formalise them to evaluate whether a solitary decision is good. 

\subsection{Implementable language}
Much like an inductive logic programming (ILP) \cite{plotkin1972} problem, we have a declarative model inferred from a set of examples. However instead of a set of propositions which refer to aspects of the world (which would once again formalise dualism), we have declarative programs and we do not need to assume those programs refer to anything because the universe is the interpreter -- all we know is that some programs return true and others false, given the state of the universe. This avoids the symbol grounding problem \cite{harnad1990}.
A subset of those declarative programs represent the physical machinery with which cognition is enacted. It follows that cognition involves only a finite number of declarative programs, because the physical constraints on cognition (memory, logic gates, time etc) tend to be finite. This finite set of declarative programs may be characterised as a language. As we have assumed the universe is the interpreter (or UTM), the meaning of this language is implemented in reality rather than interpreted by a human mind or CPU architecture. Related research described a similar notion, and called it an implementable language \cite{bennett2021}. We begin by formally defining reality, and then this language within reality, and then a task using the implementable language. 

\subsection{A note on intelligence} 
  \label{intelligence}
We have translated ``the ability to satisfy goals in a wide range of environments" as ``the ability to complete a wide range of tasks". However, Chollet argues that a better definition of intelligence is ``a measure of the ability to generalise" -- of how little information one requires to attain skill at a task \cite{chollet2020,chollet2019}. We agree. Intuitively, one's ability to generalise is one's ability learn and adapt. A greater ability to generalise entails the ability to complete a wider range of tasks (i.e. maximising the ability to generalise will maximise intelligence according to our earlier definition).

There are two reasons we prefer to describe intelligence as the ability to generalise.
First, the ability to generalise entails not only learning and adaptation, but has formed the basis of philosophical and psychological research explaining intent, the meaning of language and the ability to ascribe purpose to what is observed \cite{bennett2021,bennettmaruyama2022b,bennettmaruyama2021b,williams2022}. A description of intelligence which explains a wider range of human behaviour is more compelling than one that explains only naive utilitarianism. 

Second, this distinction is notable because AIXI's ability to satisfy goals is a consequence of its ability to generalise. In theory, the ``ability to achieve goals in a wide range of environments" \textit{could} be obtained by hard-coding action sequences, one satisfying goals in each of a fixed but wide range of environments. Given enough memory and foreknowledge any skill can be mimicked with a lookup table, much like the language computer in Searle's Chinese Room argument \cite{searle1980,dowe1998}. Thus a knowledgeable but maladaptive agent may achieve goals in a fixed range of environments. However, if we add just one environment to which the maladaptive agent's existing skills and knowledge do not apply that agent will fail. In the absence of foreknowledge, skill must be learned. AIXI is not given foreknowledge, so its ability to achieve goals in a wide range of environments must be a consequence solely of its ability to model each environment and predict the result of its actions given limited information about that environment. As discussed earlier, that ability is the result of AIXI's universal prior. 
Ergo, AIXI's particular ``ability to achieve goals in a wide range of environments" entails the ability to generalise. 

\section{Mathematical Definitions and Proofs}

\label{definitions}

To aid understanding, informal explanations accompany some of the mathematical definitions below.

\begin{remark}[Intelligence] The ability to complete a wide range of tasks, and (equivalently) the ability to generalise \cite{chollet2019}. 
\end{remark}

\begin{remark}[Artificial General Intelligence]\label{rm2} An agent that maximises intelligence (optimal across a wide range of tasks). 
\end{remark}
 
\begin{definition}[Representation of Reality]\label{d1}
A set ${H}$ containing the objective totality of each possible state of a reality, where: 

\begin{itemize}{ 
    \item  $\Omega$ is the set of all states. \footnote{The things themselves, rather than objective totalities thereof.}
    \item A declarative \textbf{program} is $f : \Omega \rightarrow \{true, false\}$, and the set of all declarative programs is $P$. A program which returns ``true" given $\omega$ is an \textbf{objective truth} about $\omega$.
    \item Given a state $\omega \in \Omega$, the \textbf{objective totality} of $\omega$ is the set of all objective truths ${h}_\omega = \{ f \in P : f(\omega) = true\}$. 
    \item ${H} = \{ j \in 2^P : \exists   \omega \in \Omega \ ( j = {h}_\omega ) \}$ is a set containing an objective totality of each state.}  
\end{itemize}
\end{definition}
\noindent{\textit{Informal Explanation.  }} Note that $P$ is equivalent to $f_3$ described in section \ref{approach}. A program is not something that needs to be perceived or interpreted as true or false by an agent, because an objective truth is a part of reality (as opposed to a mental representation of an aspect of reality). It may be understood as code of which the universe is the interpreter. If one prefers semiotic descriptions \cite{taniguchi2019}, then $p \in P$ might be characterised as a dyadic symbol that refers to itself -- it is its own referent. Any of these characterisations are fine, because they all avoid the symbol grounding problem \cite{harnad1990} and subjective interpretation. Each ${h}_\omega$ contains every objective truth about what is, what was and what will be given a state of reality $\omega$. There is a one to one correspondence between ${H}$ and $\Omega$. From the perspective of a decision, there is only one state in which that decision is made. ${H}$ contains the objective totality of each state in which decisions might be made. 
At the risk of confusing matters, for the sake of explanation each $\omega \in \Omega$ may be regarded by those comfortable with:
\begin{itemize}
    \item \textit{Reinforcement learning (AIXI)} as a sequence $\langle o_1 a_1, ... o_{t-1} a_{t-1}, o_t a_t \rangle$ of actions and observations up to time $t$, together with the associated reward.
    \item \textit{Supervised learning} as a pair $\langle input, output \rangle$.
\end{itemize}

\begin{definition}[Implementable Language] A triple $\mathcal{L} = \langle {H}, {V}, L \rangle$, where:
\begin{itemize}{ 
    \item ${H}$ is the set of \textbf{objective totalities}, each representing a state of reality. 
    \item ${V} \subset \bigcup\limits_{{h} \in {H}} {h}$ is a finite set, named the \textbf{vocabulary}.
    \item $L = \{ l \in 2^{V} : \exists {h} \in {H} \ (l \subseteq {h}) \}$ is everything that can be expressed in the language, the elements of which are \textbf{statements}. }
\end{itemize}
\end{definition}
    
\noindent{\textit{Informal Explanation.  }} ${V}$ is a finite set of declarative programs. These declarative programs are the circuitry with which cognition is enacted; including all sensors, actuators, memory etc. Because ${V}$ is finite, $L$ and every member $l \in L$ are also finite. 
A true statement $l \in L$ is any statement such that $l \subset {h}_\omega$. For each state of reality $\omega \in \Omega$, there exists an objective totality ${h}_\omega \in {H}$. Each statement must be true in at least one state of reality, because reality by definition does not permit implementation of the impossible. 

\begin{definition}[Extensions]
Given statement ${a} \in L$, the extension $Z_{a}$ of ${a}$ is $Z_{a} = \{{b} \in L : {a} \subseteq {b} \}$. Given a set ${A} \subseteq L$, the extension $Z_{A}$ of ${A}$ is $Z_{A} = \bigcup\limits_{{a} \in {A}} Z_{a}$. 
\end{definition}
\begin{remark}[Notation] To simplify presentation we'll use the following conventions for notation, except where stated otherwise:
\begin{itemize}{ 
    \item $\textbf{Statements}$ are denoted by ${s}, {d}, {m}, {z}, {c}$. 
    \item $\textbf{Sets of statements}$ are denoted by ${S}, {D}, {M}, {Z}$.
    \item The $\bf{extension}$ of an object is denoted by the capital letter Z with the object subscripted. For example the extension of ${s}$ is denoted $Z_{s}$, and the extension of ${S}$ is $Z_{S}$. }
\end{itemize}
\end{remark}

\begin{definition}[Task]\label{d3}  A task is a triple $\mathcal{T} = \langle {S}, {D}, {M} \rangle$ where:

\begin{itemize}{ 
    \item ${S} \subset L$ is a set of statements called \textbf{situations}, where the extension $Z_{S}$ of ${S}$ is the set of all \textbf{possible decisions} which can be made in those situations. 
    \item ${D}  \subset Z_{S}$ is the set of \textbf{correct decisions} for this task (goal satisficing). \footnote{Note that each ${d} \in {D}$ is a superset of a member of ${S}$. 
    ${S}$ may be understood as a set of inputs, and ${D}$ as the set of all unions of input and output which are correct.}
    \item ${M} \subset L$ is the set of all valid \textbf{models} (sets of rules) for the task, where $$
    {M} = \{{m} \in L : {Z}_{S} \cap Z_{m} \equiv {D}, \forall z \in Z_{m} \ (z \subseteq \bigcup\limits_{{d} \in {D}} {d})\}$$}
\end{itemize}
\end{definition}

\noindent{\textit{Informal Explanation.  }} The above definition is intended to encompass anything and everything one might call a task, including all sorts of ML and AI problems.
For example, those familiar with:
\begin{itemize}
    \item \textit{AI planning and constraint satisfaction problems} may interpret ${s} \in {S}$ as a partial assignment of values to variables, and $Z_{s}$ as the set of all possible complete and partial assignments given ${s}$. Each ${m} \in {M}$ is a set of constraints, and ${D}  \cap Z_{s}$ is those assignments which satisfy ${m}$ given ${s}$. As a planning problem may be represented as a constraint satisfaction problem \cite{kautz92}, this description may also be interpreted as a planning problem in which ${m}$ is a goal (and constraints that define acceptable plans), ${s} \in {S}$ is an initial state and ${d} \in {D}$ is a plan.
    \item \textit{Machine learning} may interpret ${S}$ as a set of inputs, and $Z_{S} = {D}  \cup \overline{{D}}$ as containing all possible examples of the form $\{ input, output \}$ such that $\{input\} \in {S}$. ${M}$ contains a description of every decision boundary sufficient to discriminate between ${d} \in {D}$ and ${d} \in \overline{{D}}$. 
\end{itemize}

\begin{definition}[How a Task is Completed] An agent is: 
\begin{enumerate}{ 
    \item presented with a situation ${s} \in {S}$, then
    \item selects $z \in Z_{s}$, called a decision.
    \item If $z \in {D}$, then the agent has made a correct decision and the task will be completed.}
\end{enumerate} 
Note that $\forall {m} \in {M} : {D}  \equiv Z_{S} \cap Z_{m}$, which means any ${m} \in {M}$ can be used to obtain the set of all correct decisions ${D}$ from ${S}$, because ${D} = \{z \in Z_{m} : \exists {s} \in {S} \ ({s} \subset z ) \}$.
\end{definition}

\begin{definition}[Probability of a Task] Let $\Gamma$ be the set of all tasks given an implementable language $\mathcal{L}$. There exists a uniform distribution over $\Gamma$ (because a task is arbitrary).
\end{definition}

\begin{definition}[Generalisation] Given two tasks $\mathcal{T}_1 = \langle {S}_1, {D}_1, {M}_1\rangle$ and $\mathcal{T}_2 = \langle {S}_2, {D}_2, {M}_2 \rangle$, a model ${m} \in {M}_1$ generalises to task $\mathcal{T}_2$ if ${m} \in {M}_2$.
An equivalent alternative definition is that ${m} \in {M}_1$ generalises to $\mathcal{T}_2$ if $Z_{{S}_2} \cap Z_{m} \equiv {D}_2$. 
Ability to generalise is the the probability that generalisation will occur given an agent's choice of models.
\end{definition}

\begin{definition}[Child-task and Parent-task] A task $\mathcal{T}_2 = \langle {S}_2, {D}_2, {M}_2 \rangle$ is a child-task of $\mathcal{T}_1 = \langle {S}_1, {D}_1, {M}_1\rangle$ if:
\begin{enumerate}{ 
    \item ${S}_2 \subset {S}_1$
    \item ${D}_2 \subseteq {D}_1$} 
\end{enumerate}
This is written as $\mathcal{T}_2 \sqsubset \mathcal{T}_1$. If $\mathcal{T}_2 \sqsubset \mathcal{T}_1$ then $\mathcal{T}_1$ is then a parent task of $\mathcal{T}_2$, and $\mathcal{T}_2$ is a child-task of $\mathcal{T}_1$.
\end{definition}

\begin{definition}[Weakness] $w : {M} \rightarrow \mathbb{N}$ is a function which accepts a model and returns the cardinality of its extension:
$$w({m}) = \lvert Z_{m} \rvert$$
This expresses the weakness of a given model (the greater the cardinality, the weaker the model).
\end{definition}

\subsection{Theoretical Results}
\label{proofs}

\begin{proposition}[Sufficiency]\label{p1}The probability that a model generalises to a parent task increases monotonically with its weakness.\end{proposition}
\begin{proof}
Let $\mathcal{T}_k = \langle {S}_k, {D}_k, {M}_k\rangle$ be a task of which the complete definition is known.
Let $\mathcal{T}_n = \langle {S}_n, {D}_n, {M}_n\rangle$ be a task to which we wish to generalise. All we know of $\mathcal{T}_n$ is that $\mathcal{T}_k \sqsubset \mathcal{T}_n$ (it is a parent task of $\mathcal{T}_k$), meaning ${D}_k \subset {D}_n$ and ${S}_k \subset {S}_n$.
\begin{enumerate}{ 
    \item The set of all decisions which may potentially be required to address the situations in ${S}_n$, and which are not required for ${S}_k$, is $\overline{Z_{{S}_k}} = A$.
    \item For any given ${m} \in {M}_k$, the set of decisions ${m}$ implies which fall outside the scope of what is required for the known task $\mathcal{T}_k$ is $\overline{Z_{{S}_k}} \cap Z_{{m}} = B$.
    \item $2^{\lvert A \rvert}$ is the number of tasks which fall outside of what it is necessary for a model of $\mathcal{T}_k$ to generalise to, and $2^{\lvert B \rvert}$ is the number of those tasks to which a given ${m} \in M_k$ does generalise.
    \item Therefore the probability that a given model ${m} \in {M}_k$ generalises to the unknown parent task $\mathcal{T}_n$ is $$p({m} \in {M}_n \mid {m} \in {M}_k, \mathcal{T}_k \sqsubset \mathcal{T}_n) = \frac{2^{\lvert B \rvert}}{2^{\lvert A \rvert}}$$}
\end{enumerate}
$p({m} \in {M}_n \mid {m} \in {M}_k, \mathcal{T}_k \sqsubset \mathcal{T}_n)$ increases monotonically with $w({m})$. 
\end{proof}

\begin{proposition}[Necessity]\label{p2}Weakness is necessary for generalisation.\end{proposition}

\begin{proof} 
Once again let $\mathcal{T}_k$ be a task of which the complete definition is known, and $\mathcal{T}_n$ be a parent task of $\mathcal{T}_k$ to which we wish to generalise, for which the complete definition is not known (meaning $\mathcal{T}_k \sqsubset \mathcal{T}_n$).
If ${m} \in {M}_k$ and $Z_{{S}_n} \cap Z_{{m}} \equiv {D}_n$ then it must be he case that ${D}_n \subseteq Z_{{m}}$. The weaker a model is, the more likely it is that ${D}_n \subseteq Z_{{m}}$. Therefore a sufficiently weak model is necessary for generalisation.
\end{proof}

\begin{remark}[Prior]\label{rm4} 
In the proof of proposition \ref{p1} we showed that the probability of a statement $l \in L$ generalising to a task $\mathcal{T}_n = \langle S_n,D_n,M_n \rangle$ given that $\mathcal{T}_k \sqsubset \mathcal{T}_n$, $\mathcal{T}_k = \langle S_k,D_k,M_k \rangle$ and $l \in M_k$ is $$p(l \in {M}_n \mid l \in {M}_k, \mathcal{T}_k \sqsubset \mathcal{T}_n) = \frac{2^{\lvert \overline{Z_{{S}_k}} \cap Z_{l} \rvert}}{2^{\lvert \overline{Z_{{S}_k}} \rvert}}$$
It follows that the probability of any statement $l \in L$ being a valid model for an arbitrary task $\mathcal{T} = \langle S,D,M \rangle$, given that tasks are uniformly distributed, is
$$p({l} \in M \mid {l} \in {L}) = \frac{2^{w({l})}}{2^{\lvert L \rvert}}$$ 
This assigns a probability to every statement $l \in L$ given an implementable language. It is a probability distribution in the sense that the probability of mutually exclusive statements sums to one\footnote{Two statements $a$ and $b$ are mutually exclusive if $a \not\in Z_b$ and $b \not\in Z_a$, which we'll write as $\mu(a,b)$. Given any $x \in L$, the set of all mutually exclusive statements is a set $K_x \subset L$ such that $x \in K_x$ and $\forall a, b \in K_x : \mu(a,b)$. It follows that $\forall x \in L, \underset{b \in K_x}{\sum} p(b) = 1$.}. This prior may be considered universal in the sense that it assigns a probability to every conceivable hypothesis (where what is conceivable depends upon the implementable language) absent any parameters or specific assumptions about the task as with AIXI's intelligence order relation \cite[def. 5.14 pp. 147]{hutter2010}\footnote{Some may object to the classification of $p$ as universal because $V$ is finite. However $p$ remains a valid prior if $V$ is allowed to be infinite, it just isn't particularly useful to worry about infinite quantities when an infinite $V$ can never be implemented in practice.}. As the vocabulary of the implementable language $V$ is finite, $L$ must also be finite, and so $p$ is computable. 
\end{remark}

\begin{remark}[Agents]\label{rm5}
As we have formalised enactive cognition, an embodied agent and environment are represented together as a task $\mathcal{T} = \langle S,D,M \rangle$ (see \ref{fig:1}). Recall from sections \ref{approach} and \ref{binary} that those decisions considered to be ``good enough" (the contents of $D$) are the result of preferences, goals or reward signals (that compel an embodied agent to act complete the task). Each model $m \in M$ expresses a different policy of \textit{how} to go about satisfying those goals, which is equivalent with regards to the contents of $S$ and $D$. Different choices of $m$ will imply the same decisions given $s \in S$, but may imply different decisions given $s \not\in S$ (recall from the introduction that AIXI's models are equivalent with respect to the past, but may differ with respect to the future -- this emulates that behaviour). 

To compare different agents or policies in the context of a task, we compare different choices of $m$. This is because $m$ determines what decisions are made in each situation (agency as illustrated in \ref{fig:1}). The optimal agent or policy is the optimal choice of $m$ given a task. To measure the ``ability to generalise", we need to measure the probability that a model will generalise.
\end{remark}

\begin{proposition}An artificial general intelligence is an agent which always selects the weakest models.\end{proposition}

\begin{proof}
An artificial general intelligence is an agent that maximises the ability to generalise (see \ref{rm2} and \ref{rm5}). Propositions \ref{p1} and \ref{p2} show that a selecting the weakest model is both necessary and sufficient to maximise the probability that a model generalises to any arbitrary parent task (see \ref{rm4}). Therefore an AGI is an agent which always selects $\mathbf{m}$ such that:
$$\mathbf{m} \in \underset{{m} \in {M}}{\arg\max} \ w({m})$$ 
\end{proof}

\begin{remark}[Optimality]\label{rm6}
For those familiar with AIXI we'll give a simple example in terms of reward maximisation and sequential timesteps, to illustrate how the weakest model is the optimal choice of model.
Assume we have a clock variable $t \in \mathbb{N}$, incremented in discrete timesteps as $t = t+1$ from $t = 1$ to $t = n$ ($n$ is some finite time horizon). The implementable language is $\langle H,V,L \rangle$ such that $\lvert L \rvert = n$.
Assume there exists a task $\mathcal{T} = \langle S,D,M \rangle$. At each timestep we are given the complete definition of a child-task $\langle S_{1:t},D_{1:t},M_{1:t} \rangle = \mathcal{T}_{1:t} \sqsubset \mathcal{T}$.
Each successive ostensive definition is associated with a value of $t$, for each of which $\lvert S_{1:t} \rvert = t$ and $\mathcal{T}_{1:t-1} \sqsubset \mathcal{T}_{1:t}$ (in other words every timestep a new example is added to our ostensive definition).
For the sake of putting this in comparable terms to AIXI, let $Q$ be the set of all functions $q : L \rightarrow \mathbb{N}$ which might be used to assign a score to a statement, where a higher score is used to indicate a statement is more plausible. We'll call $q \in Q$ a proxy for intelligence. For example, two such functions are weakness and inverse description length. 
A proxy $q \in Q$ is chosen, and at each time step one model $m_{1:t}$ is selected such that $$m_{1:t} \in \underset{{m} \in {M}_{1:t}}{\arg\max} \ q({m}) \quad \text{  and  } \quad p(m_{1:t}) = \frac{1}{\lvert \underset{{m} \in {M}_{1:t}}{\arg\max} \ q({m}) \rvert}$$
If the chosen $m_{1:t} \in M$, then generalisation occurs (meaning the agent has understood the task) and no further choices of model are necessary. 
Let $t=g$ be the timestep at which $m_{1:t}$ is chosen such that $m_{1:t} \in M$. For simplicity, reward is computed as $$r = \underset{t=1}{\overset{n}{\sum}} f(t) \quad \text{  where  } \quad f (t) = \begin{cases} 1 & t \ge g \\ 0 & t < g \end{cases}$$
meaning the earlier the timestep $t=g$ at which generalisation occurs, the greater the reward (i.e. this is a simple way of applying a discount rate to total reward).

$$\mathbb{E}[r] = \underset{t=1}{\overset{n}{\sum}} \ p(t \ge g) $$
$$p(t \ge g) = 1 - p(t < g)$$
$$p(t < g) = \underset{k=1}{\overset{t}{\prod}} p(m_{1:k} \not\in M \mid m_{1:k} \in \underset{{m} \in {M}_{1:t}}{\arg\max} \ q({m}))$$

$\mathbb{E}[r]$ hinges upon the chosen proxy $q \in Q$, so the question ``which agent is optimal" now amounts to ``which proxy $q$ maximises $\mathbb{E}[r]$".
From propositions \ref{p1} and \ref{p2} we have that maximising $w(m)$ the weakness $w \in Q$ of the model $m_{1:t}$ is both necessary and sufficient to maximise the probability of generalisation $p(m \in {M} \mid m \in {M}_{1:t}, \mathcal{T}_{1:t} \sqsubset \mathcal{T})$. It follows that $p(t < g)$ is minimised and $\mathbb{E}[r]$ maximised by choosing weakness as the proxy for intelligence\footnote{Note that for consistency with the rest of this paper we used ostensive definitions containing only positive examples (examples of correct decisions). However, a definition of models constructed from both positive and negative (as in incorrect decisions) examples is given in the appendix, to which this result applies as well.}. In other words: $$w \in \underset{q \in Q}{\arg\max} \ \mathbb{E}[r]$$
\end{remark}

We have shown that the weakest model is an optimal choice of model to maximise the probability of generalisation (or total expected reward as in \ref{rm6}). It follows that weakness is at least as good a proxy for intelligence as compression. However, is compression necessary to maximise the probability of generalisation? If not, then it is not as good a proxy as weakness (at least in the context of an enactive model). A direct comparison between KC and weakness is impossible, because KC would require reintroducing dualism in the form of code that needs to be interpreted, which would reintroduce the problem of subjective performance. 
The closest analogue that would not cause this problem is minimum description length (MDL), which we can define as a model  
$$c_{mdl} \in \underset{{m} \in {M}}{\arg\min} \ \lvert {m} \rvert$$
This mimics the dominant computable means of assessing model plausibility given only a description in a declarative language \cite{baker_2016,budhathokivreeken2018,evans_2020a,evans_2020b,evans_2021b}. We will compare $c_{mdl}$ with a maximum weakness model $c_w$.

\begin{remark}
Just to clarify terms we're about to use, $a$ is a ``better proxy" for $b$ than $c$ if $b \rightarrow a$ and $b \not\rightarrow c$. By ``enactive context" we mean the model we have defined over the course of this paper, of cognition enacted to complete a task, that addresses the problem of subjective performance introduced by having a UTM between the agent and environment.
\end{remark}

\begin{proposition}\label{p4} In the enactive context, weakness is a better proxy for intelligence than description length.\end{proposition}

\begin{proof}
In proposition \ref{p1} we proved that maximising weakness is necessary to maximise the probability of generalisation. It follows that either minimising description length maximises weakness, or minimising description length is not necessary to maximise the probability of generalisation. Assume the former, and we'll construct a counterexample with an implementable language $\mathcal{L} = \langle {H}, {V}, L \rangle$ where
$L = \{   
    \{a,b,c,d,j,k,z\}, 
    \{e,b,c,d,k\},
    \{a,f,c,d,j\},   \\
    \{e,b,g,d,j,k,z\},  
    \{a,f,c,h,j,k\},
    \{e,f,g,h,j,k\} 
    \}$ 
and a task $\mathcal{T} = \langle {S}, {D}, {M} \rangle$ where
\begin{itemize}{ 
    \item ${S} = \{   \{a,b\}, \{e,b\}
    \}$
    \item ${D} = \{   \{a,b,c,d,j,k,z\}, \{e,b,g,d,j,k,z\}
    \}$
    \item ${M} = \{\{z\}, \{j, k \} \}$}
\end{itemize}
The weakest model is $c_w=\{j,k\}$ and the minimum description length model is $c_{mdl}=\{z\}$. This demonstrates the minimising description length does not always maximise weakness. As weakness is necessary to maximise the probability of generalisation, minimising description length is not. Hence weakness is a better proxy for intelligence than description length in the enactive context.
\end{proof}

\section{Experiments}
\label{experiments}
Proposition \ref{p4} was also verified experimentally. Included with this paper is a Python script to perform these experiments using PyTorch with CUDA, SymPy and $A^*$ \cite{paszke2019,kirk2007,meurer2017,hart1968} (see commented code and appendix for details). 
In these experiments, a toy program computes models to 8-bit string prediction tasks. An experiment must test a falsifiable hypothesis, which in this case is ``weakness is a better proxy for intelligence than compression". 

\subsection{Methods}
\label{methods}
To specify tasks with which the experiments would be conducted, we needed an implementable language to describe simple 8-bit string prediction problems. Hence there were 256 states, one for every possible 8-bit string. The possible statements were then all the expressions regarding those $8$ bits that could be written in propositional logic (the simple connectives $\lnot$, $\land$ and $\lor$ needed to perform binary arithmetic -- a written example of how propositional logic can be used in an implementable language is also included in the appendix). For efficiency, these logical expressions were implemented as either PyTorch tensors or SymPy expressions in different parts of the program, and coverted back and forth depending on what was convenient (basic set and logical operations on these propositional tensor representations were implemented for the same reason).
A task was specified by choosing ${D}  \subset L$ such that all ${d} \in {D}$ conformed to the rules of either binary addition or multiplication with 4-bits of input, followed by 4-bits of output. The experiments were made up of trials. 
The parameters of each trial were ``operation" (a function), and an even integer ``number\_of\_trials" between 4 and 14 which determined the cardinality of the set ${D}_k$ (defined below).
Each trial was divided into training and testing phases. The training phase proceeded as follows:
\begin{enumerate}
    \item A task $\mathcal{T}_n$ was generated: 
    \begin{enumerate}
        \item First, every possible 4-bit input for the chosen binary operation was used to generate an 8-bit string. These 16 strings then formed ${D}_n$.
        \item A bit between 0 and 7 was then chosen, and ${S}_n$ created by cloning ${D}_n$ and deleting the chosen bit from every string (meaning ${S}_n$ was composed of 16 different 7-bit strings, each of which could be found in an 8-bit string in ${D}_n$).
    \end{enumerate} 
    \item A child-task $\mathcal{T}_k =  \langle {S}_k, {D}_k, {M}_k  \rangle$ was sampled from the parent task $\mathcal{T}_n$. Recall, $\lvert {D}_k \rvert$ was determined as a parameter of the trial.
    \item From $\mathcal{T}_k$ two models (rulesets) were then generated; a weakest $c_w$, and a MDL $c_{mdl}$. 
\end{enumerate}
For each model $c \in \{c_w, c_{mdl}\}$, the testing phase was as follows:
\begin{enumerate}
    \item The extension $Z_c$ of $c$ was then generated.
    \item A prediction ${D}_{recon}$ was then constructed s.t. ${D}_{recon} = \{z \in Z_c : \exists {s} \in {S}_n \ ({s} \subset z ) \}$.
    \item ${D}_{recon}$ was then compared to the ground truth ${D}_n$, and results recorded. 
\end{enumerate}
Between $75$ and $256$ trials were run for each value of the parameter $\lvert {D}_k \rvert$. Fewer trials were run for larger values of $\lvert {D}_k \rvert$ as these took longer to process. The results of these trails were then averaged for each value of $\lvert {D}_k \rvert$.

\subsection{Results}
\label{experimentalresults}
\begin{remark}[Rate at which models generalised completely] Generalisation was deemed to have occurred where ${D}_{recon} = {D}_n$. The number of trials in which generalisation occurred was measured, and divided by $n$ to obtain the rate of generalisation for each model $c_w$ and $c_{mdl}$. Error was computed as a Wald 95$\%$ confidence interval.\end{remark}
\begin{remark}[Average extent to which models generalised] Even where ${D}_{recon} \neq {D}_n$, the extent to which models generalised could be ascertained. $\frac{\lvert {D}_{recon} \cap {D}_n\rvert}{\lvert {D}_n\rvert}$ was measured and averaged for each value of $\lvert {D}_k\rvert$, and the standard error computed.\end{remark}

These results (displayed in tables \ref{table1}, \ref{table2}, \ref{table3} and \ref{table4}) demonstrate that weakness is a significantly better proxy for intelligence than compression. There was no case in which $c_{mdl}$ outperformed $c_w$. 
The generalisation rate for $c_w$ greatly exceeded that of $c_{mdl}$, and the extent of generalisation was also far greater for weakness than description length.

\begin{table}[t]
  \caption{$c_w$ results for binary addition.}
  \label{table1}
  \centering
  \begin{tabular}{lllll}
    $\lvert {D}_k \rvert$     & Rate & $\pm 95 \%$    & AvgExt & StdErr  \\
    6 & .11 & .039 & .75 & .008     \\
    10 & .27 & .064 & .91 & .006       \\
    14 & .68 & .106 & .98 & .005   
  \end{tabular}
\end{table}
\begin{table}[t]
  \caption{$c_{mdl}$ results for binary addition.}
  \label{table2}
  \centering
  \begin{tabular}{lllll}
    $\lvert {D}_k \rvert$     & Rate & $\pm 95 \%$    & AvgExt & StdErr  \\
    6 & .10 & .037 & .48 & .012    \\
    10 &  .13 & .048 & .69 & .009     \\
    14 &  .24 & .097 & .91 & .006  
  \end{tabular}
\end{table}
\begin{table}[t]
  \caption{$c_w$ results for  binary multiplication.}
  \label{table3}
  \centering
  \begin{tabular}{lllll}
    $\lvert {D}_k \rvert$     & Rate & $\pm 95 \%$    & AvgExt & StdErr  \\
    6 & .05  & .026 & .74 & .009     \\
    10 & .16 & .045 & .86 & .006       \\
    14 & .46 & .061 & .96 & .003    
  \end{tabular}
\end{table}
\begin{table}[t]
  \caption{$c_{mdl}$ results for  binary multiplication.}
  \label{table4}
  \centering
  \begin{tabular}{lllll}
    $\lvert {D}_k \rvert$     & Rate & $\pm 95 \%$    & AvgExt & StdErr  \\
    6  & .01  & .011 & .58 & .011    \\
    10  & .08 & .034 & .78 & .008     \\
    14  & .21 & .050 & .93 & .003  
  \end{tabular}
\end{table}

\section{Discussion and Limitations}
\label{conclusion}
If intelligence is a measure of the ability to generalise, then it is maximised by constructing the weakest models.
Related philosophy argues that the weakest models are the purpose of a task \cite{bennett2021}. It follows that intelligence is characterised by the ability to learn the ends (goals, rules etc), not just the means (decisions, actions, labels etc). Furthermore, the experimental results and proposition \ref{p4} undermine orthodox positions on the relationship between compression and intelligence \cite{chaitin2006,orallo2010,legg2011}, as we show that compression is neither necessary nor sufficient for intelligence (assuming cognition is indeed enactive -- which the evidence suggests it is \cite{ward2017}).

\subsection{Improving the performance of incumbent systems}
\label{connectionist}
The advantage of defining an arbitrary task is that it is general enough to encompass existing formalisms and approaches. For example, definition 3 briefly describes how supervised learning, planning and constraint satisfaction may all be described as a tasks\footnote{A detailed example of a simple regression problem represented as a task is given in the appendix.}. As such, these results can and should be applied in the context of existing tools and formulations. According to Sutton, two general purpose methods that seem to scale arbitrarily are \textit{search} and \textit{learning} (namely function approximation) \cite{sutton2019}. We have described intelligence in terms of search and declarative programs, but the concepts are as applicable to curve fitting, and to connectionist as to symbolic interpretations. The formalism may be described with category theory \cite{eilenberg1945,marquis2021}, and with imperative programs (recall that there exists an isomorphism between imperative and declarative programs \cite{howard1980}). In practical terms, the rules determining correctness can be modelled with neural networks by training a classifier $o : Z_{S} \rightarrow [0, 1]$ such that $o(l) \approx 1$ if $l \in {D}$, and $o(l) \approx 0$ otherwise. $o$ behaves as a task model, the weakness of which can then be maximised. $o$ can be used to train another network $n : {S} \rightarrow {D}$ which outputs correct decisions. $o$ learns the desired end of the task, while $n$ learns means by which it can be completed. What an architecture is suitable to describe varies according to the Chomsky hierarchy \cite{deletang2022}. Using one architecture as opposed to another is equivalent to including different sorts of declarative programs in ${V}$. An upcoming paper includes an implementation of this, along with results for declarative task models constructed from positive and negative examples (see appendix). 

\subsection{The utility of intelligence and The Apperception Engine}
\label{bias}
The vocabulary ${V}$ is meant to represent the limitations of the sensorimotor system or circuitry with which cognition is enacted. An agent that selects the weakest rulesets is optimal given any ${V}$, but a choice of finite ${V}$ restricts what sort of tasks and rulesets exist. This is of practical significance because if ${V}$ does not contain the declarative programs capable of describing what the agent is compelled to achieve (by primitives of congition such as hunger, attraction etc), then the ability to identify weak rulesets (intelligence) may become useless. 

\subsubsection{Computability} $$\epsilon = \underset{{m} \in {M}}{\arg\max} \ (w({m}) - \lvert {D}  \rvert)$$ expresses how useful intelligence is in the context of a task $\langle {S},{D},{M} \rangle$. It may be that $\epsilon \rightarrow 0$ because ${D}$ is a complete description of every decision that is correct according to some computable function $f : {S} \rightarrow 2^{D}$ (after all, there is no point being intelligent if one has a lookup table of all correct decisions). However, it is also the case that $\epsilon \rightarrow 0$ if there are no declarative programs in ${V}$ capable of expressing whatever computable relationship might exist. Were we to delete declarative programs from ${V}$, ``correct" would go from deterministic to stochastic to incomputable as $\epsilon \rightarrow 0$ (i.e. there is also no point being intelligent if a problem is beyond what it is possible to comprehend using the sensorimotor circuitry ${V}$ with which one is equipped).

\subsubsection{Complexity} Conversely, if ${V}$ contains too many declarative programs, then identifying the weakest rulesets will become intractable, even if ${V}$ does contain declarative programs that describe what is ``correct". Worst case, the number of statements that might be searched to find rulesets is $\lvert Z_{S} \rvert$, which has an upper bound of $\lvert L \rvert$, which itself has an upper bound of $2^{\lvert {V} \rvert}$, meaning ${V}$ must be small if finding the weakest rulesets is to be a tractable problem.

\subsubsection{Inductive bias} Computability and tractability are not mutually exclusive. Given a set of tasks, a ``good" inductive bias is ${V}$ containing only declarative programs needed to describe exactly what correct is for those tasks. Deepmind's Apperception Engine \cite{evans_2020a,evans_2020b,evans_2021b} demonstrates how a curated vocabulary can provide such domain specific inductive biases. It contains only rules (declarative programs) needed to describe spatial relations over time. Coincidentally, our results suggest the reason The Apperception Engine can generalise so well is because it permits only universally quantified rules, meaning if it were applied in the context of a task $\mathcal{T} = \langle {S},{D},{M} \rangle$, it would only consider models ${m} \in {M}$ such that $p({m}) = 1$. This has the effect of maximising weakness for the subset of possible tasks the vocabulary it employs can describe.

\subsection{Scale is not all you need}
The recent success of search and learning at scale \cite{sutton2019} suggests that all that is required of future AI research is to increase the scale of existing methods \cite{caballero2022}. This argument amounts to the declaration that the result of any inductive bias can be learned with enough scale (that inductive bias is just foreknowledge). Acknowledging that the debate continues regarding this point \cite{sorscher2022,deletang2022}, and that scale is all you need dismisses the cost of scale, lets assume for the sake of argument that scale is a viable approach and inductive biases are unnecessary. By fitting a curve, a neural network is approximating a model a task (albeit usually in an imperative rather than declarative form). There are many functions that fit any given curve, in much the same way there may be many valid models given a task. Just because a model is valid, does not mean it is the best choice. Intelligence is a measure of how much information the neural network requires to arrive at a correct model. We can optimise for intelligence by maximising the weakness of hypothesised models. If there is an equivalent task model given a function (see \ref{connectionist}), then the weakness of that model can be computed. As it stands, back-propagation does not optimise for weakness (even if we use a neural network to model a task in a declarative fashion, as discussed in \ref{connectionist}, additional modifications are necessary). Yes, any task can be learned by a neural network with enough scale, but that does not mean the neural network possesses intelligence. Indeed, the amount of data required to train today's models suggests the opposite. If one desire's intelligence and not mere mimicry, then scale is not all you need. You must also optimise for weakness.

\subsection{Future research on applications}
The application of these ideas at scale presents several challenges which future research may address.
First is that ${V}$ should be designed so that $\lvert {V} \rvert$ is minimised (to minimise complexity) and $\epsilon = \underset{{m} \in {M}}{\arg\max} (w({m}) - \lvert {D}\rvert)$ maximised. 
This is regardless of whether an approach is symbolic (as presented here) or connectionist (employing neural networks as mentioned earlier). 
Second, the problem of attributing task completion to decisions in order to create ostensive definitions is non-trivial, and results may vary wildly depending on how this is implemented. 
Third, attribution requires a measure of success. Success is determined by preferences, which must resemble human primitives of cognition (hunger, attraction etc) if an agent is to communicate and co-operate with humans \cite{bennett2022c}. Finally, a means of estimating the weakness of imperative and approximate models (i.e. the dominant approach to machine learning) would be useful to improve performance in such cases.



\appendices

\section{Frequently Asked Questions}
\label{faq}
\subsection{How would you apply this to solve a typical regression problem?}

\IEEEPARstart{S}{ay we} have a finite set of input values $X \subset \mathbb{R}$ and output values $Y \subset \mathbb{R}$ \footnote{We assume finite $X$ and $Y$ for practical reasons, for example that the real numbers we can represent as floating point values in a computer are constrained by the number of bits used.}, and $g : X \rightarrow Y$ be a function we wish to model. $G = \{(x,y) \in X \times Y : g(x) = y\}$, and we call $G$ the ground truth. Let $Train \subset G$ be a training set, and $Test \subset G$ be a test set. We are given $Train$ and $Test$, and our goal is to infer $G$. Typically, machine learning could be used to obtain an approximation of $G$ from $Train$ by doing the following:
\begin{enumerate}
    \item Fit a function $f$ (e.g. a neural net) such that $\forall (x,y) \in Train \left( f(x) \approx y \right)$\footnote{$a \approx b$ just means that there exists a very small number $c$ and a measure of distance $d$ such that $d(a,b)<c$, meaning the distance $d(a,b)$ between $a$ and $b$ is less than $c$}.
    \item Measure test accuracy as $\frac{\lvert \{(x,y) \in Test : y \approx f(x) \} \rvert}{\lvert Test \rvert}$.
    \item If accuracy good enough, use $f$ to make predictions $P = \{(x,y) \in X \times Y : f(x) = y\}$ and hope that $\frac{\lvert \{(x,y) \in P : y \approx g(x) \} \rvert}{\lvert P \rvert} \approx 1$.
\end{enumerate} 
This is how we would solve the problem using machine learning normally. Now let us represent this as a task and solve it that way.

\begin{enumerate}
    \item We start by creating an implementable language. 
    \begin{enumerate}
        \item Begin by defining the vocabulary $V$ of the implementable language $\mathcal{L} = \langle H, V, L \rangle$\footnote{The choice of $V$ permit an isomorphism between $X \times Y$ and $2^V$}. We need to create $V$ first and then $L$, because we cannot actually create $H$ \footnote{$H$ is the set of objective totalities of states of the universe, which may be infinite and are certainly impractical to represent. Subsequently we must obtain $L$ from directly $V$ using the structure inherent in the values of $X$ and $Y$ (e.g. $X$ and $Y$ represent real numbers in different parts of memory in a computer, not the same part, and so we can create statements in $L$ describing values from both without creating problems).}.
        \item To obtain $L$, we must first write a program $converter : X \cup Y \rightarrow 2^V$ which converts members of $X$ and $Y$ into sets of declarative programs, and another program $converter^{-1} : 2^V \rightarrow X \cup Y$ which reverses that process (meaning an isomorphism between $X \cup Y$ and $2^V$). 
        \item Next we define $pair\_converter : X \times Y \rightarrow 2^V$ such that $\forall (x,y) \in X \times Y$, $pair\_converter((x,y)) = converter(x) \cup converter(y)$ and likewise the inverse $pair\_converter^{-1}((converter(x) \cup converter(y))) = (x,y)$.
        \item A set $Q$ can be created as $Q = \{v \in 2^V : \exists (x,y) \in X \times Y \left( pair\_converter((x,y)) = v \right) \}$. We can then use $Q$ to create $L$ as each member of $Q$ must be a subset of an objective totality $h \in H$ (even though we haven't needed to explicitly define $H$), meaning $L = \{ l \in 2^{V} : \exists {v} \in {Q} \ (l \subseteq v) \}$.
    \end{enumerate}
    \item Now we can use $converter$ and $pair\_converter$ to define a task $\langle S, D, M \rangle$. 
    \begin{enumerate}
        \item First we compute $S = \{s \in L : \exists (x,y) \in Train \left( converter(x) = s  \right) \}$.
        \item Second we create the set of correct decisions $D = \{d \in L : \exists (x,y) \in Train \left( pair\_converter((x,y)) = d  \right) \}$.
        \item Finally to create $M \subset L$ by excluding members of $L$ according to the definition: $${M} = \{{m} \in L : {Z}_{S} \cap Z_{m} \equiv {D}, \forall z \in Z_{m} \ (z \subseteq \bigcup\limits_{{d} \in {D}} {d})\}$$
    \end{enumerate}
    \item We now have a set of models $M$ and situations $S$ which we can treat as constraints, and can define a program $search : S, M \rightarrow D$ which, given any situation and model, returns a decision in $D$. We can now measure the accuracy of a given model $m \in M$.
    \begin{enumerate}
        \item First we compute $S_{Test} = \{s \in L : \exists (x,y) \in Test \left( converter(x) = s  \right) \}$.
        \item Compute $D_{Test} = \{l \in L : \exists s \in S_{Test} \left( search(s,m) = l \right)\}$.
        \item Convert to real numbers by computing: $$P = \{(x,y) \in X \times Y : \gamma{(x,y)} \}$$ where $\gamma{(x,y)}$ means $$\exists d \in D_{Test} \left( pair\_converter^{-1}(d) = (x,y) \right)$$
        \item Measure accuracy as: $$\frac{\lvert \{(x,y) \in Test : \exists (a,b) \in P \left( (x,y) \approx (a,b) \right) \} \rvert}{\lvert Test \rvert}$$
    \end{enumerate}
    \item Assuming test accuracy is acceptable, we can use this to predict the ground truth $G$. 
    \begin{enumerate}
        \item Compute $S_X = \{l \in L : \exists x \in X \left( converter(x) = l  \right)  \}$, which is the set of all situations in which we need to make a decision.
        \item Choose a model $m \in M$.
        \item Compute $D_{Predicted} = \{l \in L : \exists s \in S_X \left( search(s,m) = l \right)\}$.
        \item Convert to real numbers by computing $$G_{Predicted} = \{(x,y) \in X \times Y : \delta(x,y) \}$$ where $\delta(x,y)$ means $$\exists d \in D_{Predicted} \left( pair\_converter^{-1}(d) = (x,y) \right)$$
    \end{enumerate}
\end{enumerate}
\subsection{Why would one bother representing regression or other problems as tasks?}
We can measure the weakness of a model in the context of a task and determine which model is most likely to generalise. We cannot say the same of typical approaches to learning (i.e. approximation by curve fitting).

\subsection{Why represent reality in this way?}
In order to overcome the problem of subjective performance identified by Leike and Hutter \cite{leike2015}, and to formalise enactive cognition.

\subsection{Where exactly is agency to be found in all of this?}
There is a formal definition of ``agent" based upon causality:
\begin{quote}
    ``Agents are systems that would adapt their policy if their actions influenced the world in a different way."\cite{kenton2022}
\end{quote}
In other words what defines an agent is that its decisions are made in service of a goal. If the results of decisions were different, they would not serve the agent's goals and so the agent would make different decisions.

As we have formalised enactive cognition, we do not define the agent separately from the environment, or an agent's goals separately from its model of the world. Instead, we have a task, part of which satisfies the above definition of agent.

Assume we have a task $\mathcal{T}_1 = \langle S_1,D_1,M_1 \rangle$. As section \ref{binary} describes, the agent is assumed to be compelled by primitives such as ``hunger". $\mathcal{T}_1$ is defined by these urges, describing an aspect of how such urges may be satiated to some degree or another deemed ``good enough" (presumably as determined by natural selection, to facilitate survival \cite{bennett2021,bennettmaruyama2021a}). If the result of the decisions in $D_1$ were different, then those decisions may not be ``good enough" to facilitate survival any more. The agent would need to make different decisions in the same situations, defining a new task $\mathcal{T}_2 = \langle S_1,D_2,M_2 \rangle$. The models also subsequently change. Models determine behaviour (what decisions are made), much as a policy would in the above definition of agent. In other words, because the models (policy) change when the effect of decisions change, a task describes agency as per the definition above. Moreover, this paper did not originate the idea of agency expressed as tasks determined by primitives of cognition. The issue has been discussed at length in philosophical texts which informed this research \cite{bennett2021,bennettmaruyama2021a}.



\subsection{What is the difference between ASI vs AGI?}
\label{optimality}
The difference between an AGI and an artificial super intelligence (ASI) is the implementable language employed. An AGI performs optimally given any task. However, what tasks are available depends upon the implementable language $\mathcal{L} = \langle {H},{V}, L \rangle$. An ASI is an AGI employing a vocabulary ${V}$ which maximises the utility of intelligence. 

$$\Lambda = \bigcup\limits_{{h} \in {H}} h $$
$$\epsilon =  \underset{{m} \in {M}}{\arg\max} \ (w({m}) - \lvert {D}  \rvert)$$
$$ \underset{{V} \in 2^{\Lambda}}{\arg\max} \ \epsilon $$

In other words, if we have a set $F$ of functions that are used to generate tasks given any implementable language, then an ASI is an AGI that uses the implementable language which maximises the utility of intelligence for the tasks in $F$.

\section{Example of a task}
\subsection{Example of an implementable language}
\begin{itemize}
    \item Assume there are 4 bits $bit_1, bit_2, bit_3$ and $bit_4$, and all members of ${H}$ include an assignment of values to those 4 bits. 
    \item ${V} = \{a,b,c,d,e,f,g,h,i,j,k,l\}$ is a subset of all logical tests which might be applied to these 4 bits:
    \begin{itemize}
        \item $a : bit_1 = 1$
        \item $b : bit_2 = 1$
        \item $c : bit_3 = 1$
        \item $d : bit_4 = 1$
        \item $e : bit_1 = 0$
        \item $f : bit_2 = 0$
        \item $g : bit_3 = 0$
        \item $h : bit_4 = 0$
        \item $i : j \land k$
        \item $j : bit_1 = bit_3$
        \item $k : bit_2 = bit_4$
        \item $l : i \lor bit_2 = 1$
    \end{itemize}
    \item $L = \{   
    \{a,b,c,d,i,j,k,l\}, 
    \{e,b,c,d,k,l\},
    \{a,f,c,d,j\},   \\
    \{e,f,c,d\},
    \{a,b,g,d,k,l\}, 
    \{e,b,g,d,i,j,k,l\}, \\
    \{a,f,g,d\}, 
    \{e,f,g,d,j\}, 
    \{a,b,c,h,j,l\}, 
    \{e,b,c,h,l\}, \\
    \{a,f,c,h,i,j,k,l\}, 
    \{e,f,c,h,k\},
    \{a,b,g,h,l\}, \\
    \{e,b,g,h,j\},
    \{a,f,g,h,k\}, 
    \{e,f,g,h,i,j,k,l\}
    \}$
\end{itemize}

\subsection{Example of a task $\mathcal{T}_1$}
\begin{itemize}
    \item ${S} = \{   \{a,b\}, \{e,b\},
    \{a,f\}, \{e,f\}
    \}$
    \item ${D} = \{   \{a,b,c,d,i,j,k,l\}, \{e,b,g,d,i,j,k,l\}, \\ \{a,f,c,h,i,j,k,l\}, \{e,f,g,h,i,j,k,l\}
    \}$
    \item ${M} = \{ \{i \} , \{j, k \} , \{i, j, k \}, \{i, l\}... \}$
\end{itemize}

\subsection{Example of a sufficient child-task $\mathcal{T}_2$ of $\mathcal{T}_1$}
\begin{itemize}
    \item ${S} = \{   \{a,b\}, \{e,b\}
    \}$
    \item ${D} = \{   \{a,b,c,d,i,j,k,l\}, \{e,b,g,d,i,j,k,l\}
    \}$
    \item ${M} = \{\{i,j,k,l\}, \{b,d,j\}, ... \}$
    \begin{itemize}
        \item Weakest (intensional) model $\mathbf{m} = \{i,j,k,l\}$
        \item Strongest (extensional) model $\mathbf{e} = \{b,d,j\}$
        \item $Z_\mathbf{m} = \{   
        \{a,b,c,d,i,j,k,l\}, 
        \{e,b,g,d,i,j,k,l\}, \\
        \{a,f,c,h,i,j,k,l\}, 
        \{e,f,g,h,i,j,k,l\}
        \}$
        \item $Z_\mathbf{e} = \{   
        \{a,b,c,d,i,j,k,l\}, 
        \{e,b,g,d,i,j,k,l\}
        \}$
    \end{itemize}
\end{itemize}

\section{Supplementary definitions}
\label{supplemental}
The formulation of tasks upon which this was based used different terms. Models were named solutions or rulesets, decisions were named responses and so on. These terms were changed so that the theory would be more understandable. Though not defined above, intensional and extensional solutions, as they are called in preceding work, may be useful for future work if defined as part of the formulation given in this paper. The theory also accounts for models inferred from both positive and negative examples (see below). 

\begin{definition}[Intensional model] $$\mathbf{m} \in \underset{{m} \in {M}}{\arg\max} \ w({m})$$
\end{definition}

\begin{definition}[Extensional model] $$\mathbf{e} \in \underset{{m} \in {M}}{\arg\min} \ w({m})$$

Intensional and extensional models provide bounds on weakness such that $$\forall {m} \in M : w(\mathbf{e}) \le w({m}) \le w(\mathbf{m})$$
$$\mathbf{e} \equiv Z_\mathbf{e} = {D}  \subseteq Z_\mathbf{m} \equiv \mathbf{m}$$
\end{definition}

\begin{definition}[Sufficient child-task or ostensive definition] A child-task $\mathcal{T}_2$ of $\mathcal{T}_1$ is sufficient if for every intensional model $\mathbf{m}_2 \in {M}_2$ there exists an intensional model $\mathbf{m}_1 \in {M}_1$ such that $\mathbf{m}_2 = \mathbf{m}_1$. In other words:
$$\underset{{m}_2 \in {M}_2}{\arg\max} \ w({m}_2) \subseteq \underset{{m}_1 \in {M}_1}{\arg\max} \ w({m}_1)$$
In other words every intensional model to $\mathcal{T}_2$ generalises to $\mathcal{T}_1$. We call $\mathcal{T}_2$ sufficient because it provides sufficient information to be absolutely certain of the rules of the parent task. $\mathcal{T}_2$ is also called a sufficient ostensive definition of $\mathcal{T}_1$, as per previous publications on the topic.
\end{definition}

\begin{definition}[Intensional model from positive and negative examples] Assume we wish to learn task $\mathcal{T}_0 = \langle {S}_0, {D}_0, {M}_0 \rangle$. We am given tasks $\mathcal{T}_p$ and $\mathcal{T}_n$ to learn from. $\mathcal{T}_p = \langle {S}, {D}_p, {M}_p \rangle$,  where ${D}_p \subset {D}_0$, is a child-task demonstrating what $\mathcal{T}_0$ is. $\mathcal{T}_n = \langle {S}, {D}_n, {M}_n\rangle$, where ${D}_n \cap {D}_0 = \emptyset$, demonstrates what $\mathcal{T}_0$ is not. To identify a model ${m} \in {M}_p$ which takes into account both positive and negative examples, we must first select a pair of models $p \in {M}_p$ and $n \in {M}_n$ such that $Z_p \cap Z_n = \emptyset$. \\
The set of all such pairs is $$P = \{(p,n) \in {M}_p \times {M}_n : Z_p \cap Z_n = \emptyset \}$$\\
The weakness $v : P \rightarrow \mathbb{N}$ of a pair $(p,n) \in P$ is $$v(p,n) = w(p) + w(n)$$\\
The intensional pairs of models is $$P_\mathbf{m} = \underset{(p,n) \in P}{\arg\max} \ v(p,n)$$\\
Finally, an intensional model is $$\mathbf{m} \in {M}_p : \exists (p,n) \in P_\mathbf{m} \ where \ p = \mathbf{m}$$ 
\end{definition}

\section{Additional Context}
\label{foundation}
This section discusses the philosophical notions that relate to the formal mathematical definitions introduced in section \ref{definitions}. 
The purpose of this section is to give some intuition as to what those definitions mean.

To describe an arbitrary task, we'll begin with the notions of intension and extension from philosophy of language.

\subsection{Semantic theories of meaning} Linguistic expressions can be treated as mathematical expressions \cite{frege1906}, the meaning of which is the set of things of which an expression is true. 
That set of things is called an \textbf{extension}.
Two expressions that share an extension are logically equivalent. However, there can be meaningful differences between logically equivalent sentences \cite{speaks2021}.
Quine demonstrated this with an example \cite{quine1986}:
\begin{enumerate}
    \item all animals with hearts have hearts
    \item all animals with hearts have kidneys
\end{enumerate}
Both expressions refer to the same set of animals, and so both expressions have the same extension. Yet the first expression is an obvious tautology, while the second reveals the potentially useful fact that all animals with hearts also have kidneys. Their content, or meaning as a human might interpret it, is different.
Another way to say two logically equivalent expressions have different content is to say they have different \textbf{intensions}.
Logically equivalent sentences share an extension, but may differ in their intension. However, it is important to note that an intension and its extension remain logically equivalent.
 
$$intension \leftrightarrow extension$$

\subsection{A semantic explanation of tasks}

For the sake of explanation, there is some abuse of notation in this section ($\circ, \subset$ and $\leftrightarrow$).

\subsection{Tasks as rules} A task is a concept like any other, with intensions and an extension \cite{bennettmaruyama2021b}. Consider the task ``play chess". 
$$play \ chess = intension \leftrightarrow extension$$
A game of chess is a sequence of moves by both players. The extension of ``play chess" is the set of all games of chess. An intension of this extension is any description of the rules sufficient to imply the extension.
Rules are like declarative programs. A sequence of moves can be described with a declarative program, and so each game is just a very specific rule that applies in only one game (itself). Therefore both intension and extension are just sets of rules.
What distinguishes intension from extension is how specific those rules are (in how many games is each rule applied?). The extension is formed of the most specific rules it is possible to identify (each rule applies in only one game), while the intension is made up of the weakest, least specific rules (each rule applies in all games)  \cite{bennett2021}. Though each set contains different rules, each set as a whole is equivalent to the other.
$$play \ chess = rules \leftrightarrow games$$
There may be many viable sets of rules which are neither intension (weakest) nor extension (most specific), formed of moderately weak rules sufficient to imply the extension. These are ``between" intension and extension (more intensional or extensional), and so intension and extension are two extremes of a domain possible sets of rules, each of which is as a whole equivalent but formed of rules of differing weakness \cite{bennett2021}. 

\subsection{Goals are biased rules} Biased rules are those that favour one party over another. The rules of chess are unbiased, but there is more to the notion of a task than unbiased rules. This is exemplified by the fact that ``play chess" is a different task than ``play chess and win"  \cite{bennett2021}. The former is a strictly legal description, while the latter endorses one party. 
Goals can also be described as sets of rules, and often are in the context of AI planning \cite{kautz92}. 
The extension of ``play chess and win" is all games in which the agent undertaking the task wins. The intensions are the unbiased rules of chess, integrated in some manner with biased rules describing the goal, the opponent's behaviour and everything else necessary to imply the extension.
$$rules_{bias} = unbiased \ rules \circ biased \ rules$$
$$wins \subset games$$
$$play \ chess \ and \ win = rules_{bias}  \leftrightarrow wins$$

\subsubsection{Learning models from ostensive definitions} Now suppose one wanted to learn the task ``play chess". One has ``understood" a task if one has inferred an intension that implies the correct extension -- after all, one who memorised the set $games$ without inferring the $rules$ would be labelled a mimic, not one who has understood.
The intensional $rules$ can be obtained from $games$, and so the set $games$ would be sufficient learning material to ensure one can learn how to ``play chess".
However, this is an unreasonable standard. A human trying to learn a task is not given its extension, but a set of examples called an ostensive definition \cite{gupta2021}.
As Russell put it ``all nominal definitions, if pushed back far enough, must lead ultimately to terms having only ostensive definitions" \cite{russell1948}. 
It is not enough to simply take a subset of an extension and call it an ostensive definition, as this would just specify a different task. This is demonstrated by the fact that $wins \subset games$, and yet $rules_{bias} \not\equiv rules$. 
Humans can convey concepts using ostensive definitions because we take context into account. If one learns from examples, then one must separate the rules of a task from the situations (inputs) to which they're applied. 
An ``example" is a demonstration of rules (a model which is a set of declarative programs) applied to a \textbf{situation} in order to arrive at a \textbf{decision}. If a task is to be described as a set of examples, then the extension is the set of all decisions, and an intension is a set of declarative rules (a model made of declarative programs) applied to situations (inputs). 
$$task = model \circ situations \leftrightarrow decisions$$
Returning to the chess example, assume we want to turn $wins$ into an ostensive definition of ``play chess". To obtain $rules$ from $wins$, we must describe the $situations$ in which $rules$ are applied to obtain $wins$, meaning
$$rules \circ situations \leftrightarrow wins$$
The smaller the set of examples one requires to infer the true underlying rules (model), the faster one learns. To infer those rules from an ostensive definition is to learn the task (the model obtained from the ostensive definition ``generalises" to the task as a whole). The models most likely to generalise are the most intensional, given the situations (inputs) \cite{bennett2021}. Therefore, the ability to generalise is maximised by choosing the weakest models.



\section*{Acknowledgment}
The following people gave useful feedback on this paper: Vincent Abbott, Matthew Aitchison, Jonathon Schwartz, David Quarel and Yoshihiro Maruyama all of the ANU, Badri Vellambi of UC and Sean Welsh. Their contributions significantly improved the clarity and rigour of this work, for which I am sincerely grateful. In particular, Vincent's suggestions on simplifying the formulation, and Matthew's suggestions regarding the communication of experimental evidence, were very much appreciated. 

\ifCLASSOPTIONcaptionsoff
  \newpage
\fi



%

\printbibliography

@article{cohen2022,
author = {Cohen, Michael K. and Hutter, Marcus and Osborne, Michael A.},
title = {Advanced artificial agents intervene in the provision of reward},
journal = {AI Magazine},
volume = {43},
number = {3},
pages = {282-293},
doi = {https://doi.org/10.1002/aaai.12064},
url = {https://onlinelibrary.wiley.com/doi/abs/10.1002/aaai.12064},
eprint = {https://onlinelibrary.wiley.com/doi/pdf/10.1002/aaai.12064},
year = {2022}
}

@misc{sutton2019, 
  author		= "Richard Sutton",
  title			= "The Bitter Lesson", 
  year			= "2002",
  note			= "\url{http://www.incompleteideas.net/IncIdeas/BitterLesson.html} (accessed Oct 20, 2022)"
}

@misc{deletang2022,
  doi = {10.48550/ARXIV.2207.02098},
  url = {https://arxiv.org/abs/2207.02098},
  author = {Delétang, Grégoire and Ruoss, Anian and Grau-Moya, Jordi and Genewein, Tim and Wenliang, Li Kevin and Catt, Elliot and Hutter, Marcus and Legg, Shane and Ortega, Pedro A.},
  title = {Neural Networks and the Chomsky Hierarchy},
  publisher = {arXiv},
  year = {2022},
  copyright = {Creative Commons Attribution 4.0 International}
}

@article{hart1968,
    author={Hart, Peter E. and Nilsson, Nils J. and Raphael, Bertram},
    journal={IEEE Transactions on Systems Science and Cybernetics}, 
    title={A Formal Basis for the Heuristic Determination of Minimum Cost Paths}, 
    year={1968},
    volume={4},
    number={2},
    pages={100-107},
    doi={10.1109/TSSC.1968.300136}
}

@InCollection{piccinini2021,
	author       =	{Piccinini, Gualtiero and Maley, Corey},
	title        =	{{Computation in Physical Systems}},
	booktitle    =	{The {Stanford} Encyclopedia of Philosophy},
	editor       =	{Edward N. Zalta},
	howpublished =	{\url{https://plato.stanford.edu/archives/sum2021/entries/computation-physicalsystems/}},
	address = {Stanford},
	year         =	{2021},
	edition      =	{{S}ummer 2021},
	publisher    =	{Metaphysics Research Lab, Stanford University}
}

@article{meurer2017,
    author = {Meurer, Aaron and Smith, Christopher and Paprocki, Mateusz and Čertík, Ondřej and Kirpichev, Sergey and Rocklin, Matthew and Kumar, AMiT and Ivanov, Sergiu and Moore, Jason and Singh, Sartaj and Rathnayake, Thilina and Vig, Sean and Granger, Brian and Muller, Richard and Bonazzi, Francesco and Gupta, Harsh and Vats, Shivam and Johansson, Fredrik and Pedregosa, Fabian and Scopatz, Anthony},
    year = {2017},
    month = {01},
    pages = {e103},
    title = {SymPy: Symbolic computing in Python},
    volume = {3},
    journal = {PeerJ Computer Science},
    doi = {10.7717/peerj-cs.103}
}

@inproceedings{kirk2007,
    author = {Kirk, David},
    title = {NVIDIA Cuda Software and Gpu Parallel Computing Architecture},
    year = {2007},
    isbn = {9781595938930},
    publisher = {Association for Computing Machinery},
    address = {New York, NY, USA},
    url = {https://doi.org/10.1145/1296907.1296909},
    doi = {10.1145/1296907.1296909},
    booktitle = {Proceedings of the 6th International Symposium on Memory Management},
    pages = {103–104},
    numpages = {2},
    location = {Montreal, Quebec, Canada},
    series = {ISMM '07}
}

@inproceedings{paszke2019,
    author = {Paszke, Adam and Gross, Sam and Massa, Francisco and Lerer, Adam and Bradbury, James and Chanan, Gregory and Killeen, Trevor and Lin, Zeming and Gimelshein, Natalia and Antiga, Luca and Desmaison, Alban and K\"{o}pf, Andreas and Yang, Edward and DeVito, Zach and Raison, Martin and Tejani, Alykhan and Chilamkurthy, Sasank and Steiner, Benoit and Fang, Lu and Bai, Junjie and Chintala, Soumith},
    title = {PyTorch: An Imperative Style, High-Performance Deep Learning Library},
    year = {2019},
    publisher = {Curran Associates Inc.},
    address = {Red Hook, NY, USA},
    booktitle = {Proceedings of the 33rd International Conference on Neural Information Processing Systems},
    articleno = {721},
    numpages = {12}
}

@article{chaitin2006,
    author = {Chaitin, G.},
    year = {2006},
    pages = {74--81},
    title = {The Limits of Reason},
    volume = {294},
    number = {3},
    journal = {Scientific American}
}

@inproceedings{dowe1998,
  title={A Non-Behavioural, Computational Extension to the Turing Test},
  author={David L. Dowe and Alan H{\'a}jek},
  year={1998}
}

@article{leike2018,
title = {On the computability of Solomonoff induction and AIXI},
journal = {Theoretical Computer Science},
volume = {716},
pages = {28-49},
year = {2018},
note = {Special Issue on ALT 2015},
issn = {0304-3975},
doi = {https://doi.org/10.1016/j.tcs.2017.11.020},
url = {https://www.sciencedirect.com/science/article/pii/S0304397517308502},
author = {Jan Leike and Marcus Hutter}
}

@article{orallo2010,
title = {Measuring universal intelligence: Towards an anytime intelligence test},
journal = {Artificial Intelligence},
volume = {174},
number = {18},
pages = {1508-1539},
year = {2010},
issn = {0004-3702},
doi = {https://doi.org/10.1016/j.artint.2010.09.006},
url = {https://www.sciencedirect.com/science/article/pii/S0004370210001554},
author = {José Hernández-Orallo and David L. Dowe},
}

@article{goertzel2014,
    author = {Ben Goertzel},
    year = {2014},
    pages = {1--48},
    title = {Artificial General Intelligence: Concept, State of the Art},
    volume = {5},
    number = {1},
    journal = {Journal of Artificial General Intelligence}
}

@InCollection{howard1980,
	author       =	{Howard, William A.},
	title        =	{{The Formulae-as-Types Notion of Construction}},
	booktitle    =	{To H.B. Curry: Essays on Combinatory Logic, Lambda Calculus and Formalism},
	editor       =	{Seldin, J.P. and Hindley, J.R.},
	year         =	{1980},
	pages      =	{479--490},
	publisher    =	{Academic Press},
	address = {Cambrdige MA}
}

@InProceedings{bennett2022c,
author="Bennett, Michael Timothy",
editor="Goertzel, Ben
and Ikl{\'e}, Matthew
and Potapov, Alexey",
title="Compression, The Fermi Paradox and Artificial Super-Intelligence",
booktitle="Artificial General Intelligence",
year="2022",
publisher="Springer International Publishing",
address="Cham",
pages="41--44",
isbn="978-3-030-93758-4"
}

@article{eilenberg1945,
    ISSN = {00029947},
    URL = {http://www.jstor.org/stable/1990284},
    author = {Samuel Eilenberg and Saunders MacLane},
    journal = {Transactions of the American Mathematical Society},
    number = {2},
    pages = {231--294},
    publisher = {American Mathematical Society},
    title = {General Theory of Natural Equivalences},
    urldate = {2022-08-05},
    volume = {58},
    year = {1945}
}

@InCollection{marquis2021,
	author       =	{Marquis, Jean-Pierre},
	title        =	{{Category Theory}},
	booktitle    =	{The {Stanford} Encyclopedia of Philosophy},
	editor       =	{Edward N. Zalta},
	address = {Stanford},
	howpublished =	{\url{https://plato.stanford.edu/archives/fall2021/entries/category-theory/}},
	year         =	{2021},
	edition      =	{{F}all 2021},
	publisher    =	{Metaphysics Research Lab, Stanford University}
}

@article{harnad1990,
    title = {The symbol grounding problem},
    journal = {Physica D: Nonlinear Phenomena},
    volume = {42},
    number = {1},
    pages = {335-346},
    year = {1990},
    issn = {0167-2789},
    doi = {https://doi.org/10.1016/0167-2789(90)90087-6},
    url = {https://www.sciencedirect.com/science/article/pii/0167278990900876},
    author = {Stevan Harnad}
}

@article{clark1998,
	journal = {Analysis},
	volume = {58},
	year = {1998},
	number = {1},
	title = {The Extended Mind},
	pages = {7--19},
	doi = {10.1093/analys/58.1.7},
	publisher = {Oxford University Press},
	author = {Andy Clark and David J. Chalmers}
}

@article{ward2017,
author = {Ward, Dave and Silverman, David and Villalobos, Mario},
year = {2017},
month = {04},
pages = {},
title = {Introduction: The Varieties of Enactivism},
volume = {36},
journal = {Topoi},
doi = {10.1007/s11245-017-9484-6}
}

@ARTICLE{taniguchi2019,
	author={Tadahiro {Taniguchi} and Emre {Ugur} and Matej {Hoffmann} and Lorenzo {Jamone} and Takayuki {Nagai} and Benjamin {Rosman} and Toshihiko {Matsuka} and Naoto {Iwahashi} and Erhan {Oztop} and Justus {Piater} and Florentin {Wörgötter}},
	journal={IEEE Transactions on Cognitive and Developmental Systems}, 
	title={Symbol Emergence in Cognitive Developmental Systems: A Survey}, 
	year={2019},
	volume={11},
	number={4},
	pages={494-516}
}

@InCollection{shapiro2021,
	author       =	{Shapiro, Lawrence and Spaulding, Shannon},
	title        =	{{Embodied Cognition}},
	booktitle    =	{The {Stanford} Encyclopedia of Philosophy},
	editor       =	{Edward N. Zalta},
	address = {Stanford},
	howpublished =	{\url{https://plato.stanford.edu/archives/win2021/entries/embodied-cognition/}},
	year         =	{2021},
	edition      =	{{W}inter 2021},
	publisher    =	{Metaphysics Research Lab, Stanford University}
}

@InCollection{howard2020,
	author       =	{Robinson, Howard},
	title        =	{{Dualism}},
	booktitle    =	{The {Stanford} Encyclopedia of Philosophy},
	editor       =	{Edward N. Zalta},
	address = {Stanford},
	howpublished =	{\url{https://plato.stanford.edu/archives/fall2020/entries/dualism/}},
	year         =	{2020},
	edition      =	{{F}all 2020},
	publisher    =	{Metaphysics Research Lab, Stanford University}
}

@InProceedings{kautz92,
    author = {Henry Kautz and Bart Selman},
    title = {Planning as satisfiability},
    booktitle = {IN ECAI-92},
    year = {1992},
    pages = {359--363},
	address = {New York},
    publisher = {Wiley}
}

@article{legg2007,
	pages = {391--444},
	doi = {10.1007/s11023-007-9079-x},
	title = {Universal Intelligence: A Definition of Machine Intelligence},
	author = {Shane Legg and Marcus Hutter},
	volume = {17},
	journal = {Minds and Machines},
	publisher = {Springer},
	number = {4},
	year = {2007}
}

@article{searle1980,
	author       =	{Searle, John},
	title        =	{{Minds, Brains, and Programs}},
	year         =	{1980},
    journal = "Behavioral and Brain Sciences",
    volume = "3",
    pages = "417-457"
}

@article{alexander2020,
  author    = {Samuel Allen Alexander},
  title     = {The Archimedean trap: Why traditional reinforcement learning will probably not yield {AGI}},
  journal   = {CoRR},
  volume    = {abs/2002.10221},
  year      = {2020},
  url       = {https://arxiv.org/abs/2002.10221},
  eprinttype = {arXiv},
  eprint    = {2002.10221},
  bibsource = {dblp computer science bibliography, https://dblp.org}
}

@article{wirth2017,
  author  = {Christian Wirth and Riad Akrour and Gerhard Neumann and Johannes F{{\"u}}rnkranz},
  title   = {A Survey of Preference-Based Reinforcement Learning Methods},
  journal = {Journal of Machine Learning Research},
  year    = {2017},
  volume  = {18},
  number  = {136},
  pages   = {1--46},
  url     = {http://jmlr.org/papers/v18/16-634.html}
}

@misc{chollet2020,
  title="Abstraction \& Reasoning in AI systems: Modern Perspectives",
  author={Chollet, F. and Mitchell, M. and Szegedy, C.},
  note={Thirty-fourth Conference on Neural Information Processing Systems},
  howpublished={\url{https://nips.cc/Conferences/2020/Schedule?showEvent=16644}},
  year={2020}
}

@misc{chollet2019,
  doi = {10.48550/ARXIV.1911.01547},
  url = {https://arxiv.org/abs/1911.01547},
  author = {Chollet, François},
  title = {On the Measure of Intelligence},
  publisher = {arXiv},
  year = {2019},
  copyright = {arXiv.org perpetual, non-exclusive license}
}

@book{vitanyi2008,
	publisher = {Springer},
	year = {2008},
	author = {Ming Li and Paul M. B. Vit\'{a}nyi},
	title = {An Introduction to Kolmogorov Complexity and its Applications (Third Edition)},
	address={New York}
}

@article{solomonoff1978,
  author={Ray Solomonoff},
  journal={IEEE Transactions on Information Theory}, 
  title={Complexity-based induction systems: Comparisons and convergence theorems}, 
  year={1978},
  volume={24},
  number={4},
  pages={422–432}
}

@inproceedings{legg2011,
  title={An Approximation of the Universal Intelligence Measure},
  author={Shane Legg and Joel Veness},
  booktitle={Algorithmic Probability and Friends},
  year={2011}
}

@InProceedings{bennett2021,
    author="Bennett, M.T.",
    editor="Goertzel, B.
    and Ikl{\'e}, M.
    and Potapov, A.",
    title="Symbol Emergence and the Solutions to Any Task",
    booktitle="Artificial General Intelligence",
    year="2022",
    publisher="Springer International Publishing",
    address="Cham",
    pages="30--40",
    isbn="978-3-030-93758-4"
}

@article{bennettmaruyama2021b,
  author={Bennett, Michael Timothy and Maruyama, Yoshihiro},
  journal={IEEE Transactions on Cognitive and Developmental Systems}, 
  title={Philosophical Specification of Empathetic Ethical Artificial Intelligence}, 
  year={2022},
  volume={14},
  number={2},
  pages={292-300}
}

@article{williams2022,
	author = {Williams, J. and Fiore, S.M. and Jentsch, F.},
	title = {Supporting Artificial Social Intelligence With Theory of Mind},
	journal = {Front. Artif. Intell.},
	year = {2022},
	url = {https://www.frontiersin.org/articles/10.3389/frai.2022.750763/full}
}

@book{hutter2010,
    author = {Hutter, Marcus},
    title = {Universal Artificial Intelligence: Sequential Decisions Based on Algorithmic Probability},
    year = {2010},
    isbn = {3642060528},
    publisher = {Springer-Verlag},
    address = {Berlin, Heidelberg}
}

@phdthesis{legg2008,
	author        = "Legg, Shane",
	title         = "Machine Super Intelligence",
	school     = "University of Lugano",
	year          = "2008"
}

@article{solomonoff_1964a,
  author={Solomonoff, R.J.},
  journal={Information and Control}, 
  title={A formal theory of inductive inference. Part I}, 
  year={1964},
  volume={7},
  number={1},
  pages={1-22}
}

@article{solomonoff_1964b,
  author={Solomonoff, R.J.},
  journal={Information and Control}, 
  title={A formal theory of inductive inference. Part II},
  year={1964},
  volume={7},
  number={2},
  pages={224-254}
}

@inproceedings{plotkin1972,
  title={Automatic Methods of Inductive Inference},
  author={Gordon D. Plotkin},
  year={1972}
}

@article{leike2015,
  author={Leike, J. and Hutter, M.},
  journal={Proceedings of The 28th Conference on Learning Theory, in Proceedings of Machine Learning Research}, 
  title={Bad Universal Priors and Notions of Optimality},
  year={2015},
  pages={1244-1259}
}

@article{frege1906,
  author={Frege, G.},
  journal={Unpublished}, 
  title={Kurze Übersicht meiner logischen Lehren? [Translation in Beaney, M.: The Frege Reader. Oxford Blackwell. pp. 299-300 (1997)]},
  year={1906}
}

@InCollection{speaks2021,
	author       =	{Speaks, J.},
	title        =	{{Theories of Meaning}},
	booktitle    =	{The {Stanford} Encyclopedia of Philosophy},
	editor       =	{Edward N. Zalta},
	address = {Stanford},
	howpublished =	{\url{https://plato.stanford.edu/archives/spr2021/entries/meaning/}},
	year         =	{2021},
	edition      =	{{S}pring 2021},
	publisher    =	{Metaphysics Research Lab, Stanford University}
}

@book{quine1986,
    ISBN = {9780674665637},
    note = {\url{http://www.jstor.org/stable/j.ctvk12scx}},
    author = {Quine, W.V.O.},
    publisher = {Harvard University Press},
    title = {Philosophy of Logic: Second Edition},
    urldate = {2022-08-04},
    pages={8-9},
	address = {Cambridge MA},
    year = {1986}
}

@misc{sorscher2022,
  doi = {10.48550/ARXIV.2206.14486},
  url = {https://arxiv.org/abs/2206.14486},
  author = {Sorscher, Ben and Geirhos, Robert and Shekhar, Shashank and Ganguli, Surya and Morcos, Ari S.},
  title = {Beyond neural scaling laws: beating power law scaling via data pruning},
  publisher = {Conference on Neural Information Processing Systems},
  year = {2022}
}

@misc{caballero2022,
  doi = {10.48550/ARXIV.2210.14891},
  url = {https://arxiv.org/abs/2210.14891},
  author = {Caballero, Ethan and Gupta, Kshitij and Rish, Irina and Krueger, David},
  title = {Broken Neural Scaling Laws},
  publisher = {arXiv},
  year = {2022},
  copyright = {Creative Commons Attribution 4.0 International}
}

@article{bennettmaruyama2021a,
  title={Intensional Artificial Intelligence: From Symbol Emergence to Explainable and Empathetic AI},
  author={Bennett, M. T. and Maruyama, Y.},
  journal={arXiv preprint arXiv:2104.11573 [cs.AI]},
  year={2021}
}

@book{russellandnorvig2021,
  author={Russell, S. and P. Norvig.},
  publisher={Pearson}, 
  title={Artificial intelligence: A modern approach, global edition 4th},
  pages={36},
  address = {London},
  year={2021}
}

@book{thompson2007,
	author = {Evan Thompson},
	title = {Mind in Life: Biology, Phenomenology, and the Sciences of Mind},
	publisher = {Harvard University Press},
	address = {Cambridge MA},
	year = {2007}
}

@InCollection{gupta2021,
	author       =	{Gupta, Anil},
	title        =	{{Definitions}},
	booktitle    =	{The {Stanford} Encyclopedia of Philosophy},
	editor       =	{Edward N. Zalta},
	address = {Stanford},
	howpublished =	{\url{https://plato.stanford.edu/archives/win2021/entries/definitions/}},
	year         =	{2021},
	edition      =	{{W}inter 2021},
	publisher    =	{Metaphysics Research Lab, Stanford University}
}

@book{russell1948,
  author={Russell, B.},
  publisher={Simon and Schuster}, 
  title={Human Knowledge: Its Scope and Limits},
  pages={242},
  address = {New York},
  year={1948}
}

@article{kolmogorov_1963,
  author={Kolmogorov, A.N.},
  journal={Sankhya: The Indian Journal of Statistics}, 
  title={On tables of random numbers},
  year={1963},
  volume={A},
  pages={369-376}
}

@InProceedings{bennettmaruyama2022b,
    author="Bennett, Michael Timothy
    and Maruyama, Yoshihiro",
    editor="Goertzel, Ben
    and Ikl{\'e}, Matthew
    and Potapov, Alexey",
    title="The Artificial Scientist: Logicist, Emergentist, and Universalist Approaches to Artificial General Intelligence",
    booktitle="Artificial General Intelligence",
    year="2022",
    publisher="Springer International Publishing",
    pages="45--54",
address="Cham"
}

@InCollection{baker_2016,
	author       =	{Baker, A.},
	title        =	{{Simplicity}},
	booktitle    =	{The {Stanford} Encyclopedia of Philosophy},
	editor       =	{Edward N. Zalta},
	address = {Stanford},
	howpublished =	{\url{https://plato.stanford.edu/archives/sum2022/entries/simplicity/}},
	year         =	{2022},
	edition      =	{{S}ummer 2022},
	publisher    =	{Metaphysics Research Lab, Stanford University}
}

@misc{kenton2022,
  doi = {10.48550/ARXIV.2208.08345},
  url = {https://arxiv.org/abs/2208.08345},
  author = {Kenton, Zachary and Kumar, Ramana and Farquhar, Sebastian and Richens, Jonathan and MacDermott, Matt and Everitt, Tom},
  title = {Discovering Agents},
  publisher = {arXiv},
  year = {2022}, 
  copyright = {arXiv.org perpetual, non-exclusive license}
}

@article{budhathokivreeken2018,
  author={Budhathoki, K. and Vreeken, J.},
  journal={Knowledge and Information Systems}, 
  title={Origo: Causal Inference by Compression},
  year={2018},
  volume={56},
  number={2},
  pages={28-307}
}

@phdthesis{evans_2020a,
  author={Evans, R.},
  school={Imperial College}, 
  title={Kant's Cognitive Architecture},
  year={2020}
}

@article{evans_2020b,
  author={Evans, R. and Sergot, M. and Stephenson, A.},
  journal={J Philos Logic}, 
  title={Formalizing Kant’s Rules},
  year={2020},
  volume={49},
  pages={613-680}
}

@article{evans_2021b,
  author={Evans, R. and Bošnjak, M. and Buesing, L. and Ellis, K. and Pfau, D. and Kohli, P. and Sergot, M.},
  journal={Artificial Intelligence}, 
  title={Making Sense of Raw Input},
  year={2021},
  volume={299}
}




%








\end{document}